\documentclass[10pt,twocolumn,letterpaper]{article}

\usepackage[pagenumbers]{wacv}      

\definecolor{oursrow}{HTML}{EAF0F8}   
\definecolor{refrow}{HTML}{F4F4F4}    
\definecolor{oraclerow}{HTML}{FBF0E6} 
\definecolor{belowref}{HTML}{A83232}  
\newcommand{\best}[1]{\textbf{#1}}
\newcommand{\second}[1]{\underline{#1}}

\newcommand{\std}[1]{\,\scriptsize$\pm$#1\normalsize}

\definecolor{wacvblue}{rgb}{0.21,0.49,0.74}
\usepackage{graphicx}
\usepackage{amsmath}
\usepackage{amssymb}
\usepackage{booktabs}
\usepackage{multirow}
\usepackage[table]{xcolor}
\usepackage{algorithm}
\usepackage{algpseudocode}
\usepackage{enumitem}
\usepackage{pifont}
\usepackage{comment}
\newcommand{\cmark}{\ding{51}}
\newcommand{\xmark}{\ding{55}}

\usepackage[pagebackref,breaklinks,colorlinks,allcolors=wacvblue]{hyperref}
\usepackage[capitalize]{cleveref}
\crefname{section}{Sec.}{Secs.}
\Crefname{section}{Section}{Sections}
\crefname{table}{Table}{Tables}
\Crefname{table}{Table}{Tables}

\usepackage{soul}
\def\wacvPaperID{449} 
\def\confName{WACV}
\def\confYear{2027}

\title{Rethinking Post-hoc Debiasing of Large Pretrained Models}

\author{
Divyaansh Mertia$^{*}$ \and
Chirag Bhutra$^{*}$ \and
Himanshu Shekhar$^{*}$ \and
Shubham Bagwari$^{*}$ \and
Pratik Mazumder$^{*}$\\
Indian Institute of Technology Jodhpur\\
{\tt\small \{p25cs0004, b23mt1012, b23mt1021, p22cs201, pratikm\}@iitj.ac.in}
\thanks{These authors contributed equally.}
}

\begin{document}
\maketitle
\begin{abstract}
Large pretrained models such as CLIP are widely used as frozen feature extractors. However, their embeddings inherit spurious correlations from training data, introducing bias into the features. A growing body of research has sought to address this issue through post hoc debiasing, leaving the backbone untouched. We discover that despite differing mechanisms, these methods require the bias feature to be identified and expressed in text. This dependency is not innocuous. In this work, we experimentally demonstrate that this requirement incurs an unforeseen penalty in certain scenarios. For example, on the UrbanCars benchmark, having two co-occurring bias attributes, the recent methods underperform a vanilla classifier on biased frozen-backbone features, and even providing ground-truth bias identities does not close the gap. We show that the pattern that such methods seek is readily observable in the frozen geometry. Specifically, cross-class training pairs placed closest together by the frozen encoder almost always share a spurious attribute, regardless of the bias type, without requiring access to the bias labels. Based on this, we propose a novel approach to identify such pairs and learn a single linear map that transforms the features to perform post-hoc debiasing. The proposed approach achieves competitive performance across all datasets without access to bias attribute annotations or any LLMs.
\end{abstract}

\section{Introduction}
\label{sec:intro}

Large pretrained models such as CLIP~\cite{clip} are widely used as frozen feature extractors. While classifiers built on the embeddings of such models reach high average accuracy on multiple tasks, they often fail on a predictable subset of inputs. The root cause of these failures stems from the data distributions at both the pretraining and downstream stages. Since foundational models are trained on massive, uncurated web corpora, their learned feature spaces inherently capture and encode various naturally occurring `spurious correlations' \cite{bias1,bias2,bias3,bias4,bias5,bias6}. Consequently, when a spurious attribute strongly co-occurs with the class label, classifiers exploit this shortcut, making the model \textit{biased} toward that attribute, e.g., with a majority of white dog and black cat images, the classifier exploits this color shortcut over complex class-defining features. Such classifiers will fail at test time, if this spurious correlation breaks.

Post-hoc debiasing methods address this issue by adapting the learned representation while keeping the pretrained backbone frozen~\cite{fairerclip,roboshot,prism}. Despite differing implementations, they all require the spurious attribute to be identified beforehand. Earlier methods relied on practitioners to specify the spurious attribute and construct descriptive text prompts~\cite{fairerclip}, while recent works use LLMs to propose candidate bias attributes based solely on class names and generate controlled scene descriptions. These are used to suppress the corresponding text-aligned direction in the embedding space~\cite{roboshot,prism}. While this shift is typically framed as a transition toward unsupervised debiasing, it is more appropriately just a change of oracle. This fundamental dependency on external attribute specification makes these methods incapable of detecting or correcting a misidentified bias.

This reliance on text-based attribute identification introduces a critical failure mode when datasets contain complex or co-occurring shortcuts. While LLM-driven methods succeed when a single spurious attribute aligns cleanly with broad world knowledge, they struggle with multiple or non-intuitive attributes. On the UrbanCars benchmark~\cite{urbancars}, which features two simultaneous spurious attributes (background and co-occurring object), methods that obtain their attribute from a language model perform worse than an undebiased linear classifier trained on the frozen embeddings. PRISM~\cite{prism} reaches only $3.23\%$ worst-group accuracy with a trained linear head, compared to $38.71\%$ for the undebiased baseline. 


One might hypothesize that this failure stems merely from imprecise LLM-generated descriptions. To test this hypothesis, we perform an oracle experiment by supplying ground-truth description of the spurious attributes to these methods (\cref{subsec:oracle}). The accuracy improves but remains below the undebiased baseline. This demonstrates that external bias identification is necessary for such methods but insufficient: suppressing multiple text-guided subspaces can remove essential class-discriminative semantics, damaging the representation more than the shortcuts it removes. Thus, prompt refinement cannot resolve this fundamental trade-off.

To overcome these limitations, we propose an alternative paradigm that operates directly on an intrinsic \emph{geometric signature} of bias rather than on bias attribute specifications. We define a sample as \emph{bias-aligned} if its spurious attribute agrees with the dominant spurious-label correlation in the training data, and \emph{bias-conflicting} if it violates it. Consequently, a cross-class pair sharing the same spurious attribute necessarily consists of one bias-aligned and one bias-conflicting sample, a configuration we term a \emph{bias-sharing pair}. We observe that in a biased frozen feature space, cross-class pairs exhibiting high cosine similarity are bias-sharing. Spurious correlation thus manifests geometrically as a localized blur along identifiable cross-class sample pairs. Detecting these pairs requires no text prompts or attribute supervision, only class labels and pairwise similarity. By pulling apart these confusable cross-class pairs while maintaining the intra-class cohesion, we learn a single metric transformation $\mathbf{W} \in \mathbb{R}^{d \times d}$ over the frozen embeddings. Because this geometric criterion relies strictly on feature similarity, it isolates and separates bias-sharing pairs agnostic to the number or nature of underlying shortcuts.

Our proposed framework requires no bias attribute annotations, language models, auxiliary biased models, or backpropagation through the backbone. Across extensive evaluations, our approach achieves $81.18\%$ worst-joint accuracy on UrbanCars (compared to $61.29\%$ for the best competing method) while simultaneously improving overall accuracy. On BAR, our method delivers state-of-the-art worst-group accuracy across all bias-conflicting ratios. Finally, on BFFHQ, our performance scales dynamically with the availability of conflicting samples, outperforming static prompt-based baselines as conflicting data increases. The contributions of this work are as follows:


\begin{itemize}[leftmargin=1.2em,itemsep=1pt,topsep=2pt]
  \item We empirically demonstrate that post-hoc debiasing methods that obtain the spurious attribute from a language model fail on multi-shortcut datasets, performing worse than an undebiased linear classifier, and that supplying ground-truth text prompts does not close the gap. 
  \item We identify an intrinsic geometric signature of spurious correlation in frozen embedding spaces where cross-class feature pairs with high similarity are overwhelmingly bias-sharing and show it can be detected purely from pairwise cosine similarity.
  \item We introduce an annotation-free post-hoc debiasing approach that does not rely on external attribute descriptions or language models.
  \item We evaluate the proposed approach across diverse multimodal and vision-only backbones  (including CLIP\cite{clip}, SigLIP\cite{siglip}, and DINOv3\cite{dinov3}), achieving state-of-the-art worst-group performance on the multi-shortcut benchmark and competitive performance across single-shortcut benchmarks.
\end{itemize}

\section{Related Work}
\label{sec:related}
There is a growing body of work that involves debiasing deep learning models. In this section, we discuss existing works that perform bias mitigation under various settings.
\\
\ 
\\
\noindent
\textbf{Robust training without group labels.} When group annotations are available, distributionally robust optimization~\cite{groupdro} and last-layer retraining on a group-balanced set~\cite{dfr} optimize worst-group accuracy directly. A larger body of work removes that requirement by inferring which samples are atypical: from the errors of a deliberately biased model~\cite{lff,jtt} then reweighting against them. These methods are effective, but they assume that the encoder is trainable, and their inference of atypical samples rests on a representation that changes as training proceeds.
\\
\ 
\\
\noindent
\textbf{Debiasing frozen vision--language embeddings.}
Various recent works debias pretrained vision-language models (VLMs) without updating their backbones, instead modifying the learned embedding space post-hoc. These methods are often described as \emph{annotation-free} because they do not require per-sample annotations of the spurious attribute. However, this does not necessarily remove the need to identify the attribute itself. In practice, most existing methods need to be told which attribute is spurious, either by the practitioner or by a large language model.

\emph{Bias attribute named by the practitioner.}
FairerCLIP~\cite{fairerclip} learns functions over both encoders in a reproducing kernel Hilbert space, minimizing dependence between the representation and a specified attribute while preserving dependence on the target. Although it does not require attribute annotations per image, the bias attribute must be explicitly named, and prompts describing its values are used.

\emph{Bias attribute named by a language model.}
RoboShot~\cite{roboshot} shifts the burden of specifying the bias attribute to an LLM, which proposes features that are helpful or harmful for a target class, which are then converted into embedding directions and used for projection. PRISM~\cite{prism} similarly uses an LLM to propose candidate attributes, then generates controlled scene descriptions in which the class and attribute vary independently and learns a projection that is invariant to the resulting attributes. Its variant, PRISM-mini, replaces the learned projection with an orthogonal projection. These methods avoid manually specifying the bias attribute, but rely on an external model to identify candidate attributes.

\emph{Bias attribute is not specified.}
Rather than identifying a spurious attribute, $\text{SEM}_i$~\cite{sem} searches for representation components that are broadly responsive rather than selectively associated with a target class. It encodes frozen embeddings with a sparse encoder and compares neuron responses on two prompt sets: 328 neutral captions spanning diverse scenes, objects, and activities, which provide a reference distribution, and ten LLM-generated paraphrases of each class name. It suppresses neurons with low query relevance, which are treated as likely spurious.
\\
\ 
\\
\noindent
\textbf{Multiple shortcuts.}
Li~\etal~\cite{urbancars} observe that mitigating one shortcut can amplify reliance on another, a failure they term \emph{whac-a-mole}, and introduce UrbanCars to expose it: two shortcuts are inserted at once, one of which is an artifact of dataset construction rather than a regularity of the visual world. The frozen-backbone methods above are evaluated almost exclusively on benchmarks carrying a single shortcut. We experimentally demonstrate in \cref{sec:analysis} that whac-a-mole persists on frozen embeddings, and is not resolved just by ground truth bias attribute specifications.

In this work, we present a post-hoc debiasing approach that performs post-hoc debiasing of the features of a frozen large pretrained model, without requiring any practitioner or language model or even the bias specifications. We also experimentally demonstrate that our proposed approach performs significantly better than existing post-hoc debiasing methods including on the UrbanCars benchmark.
\begin{table}[t]
\centering
\small
\setlength{\tabcolsep}{4pt}
\renewcommand{\arraystretch}{1.15}
\begin{tabular}{l l c c}
\toprule
\textbf{Method} & \textbf{Venue} & \textbf{Bias Info} & \textbf{Text-Agnostic} \\
\midrule
\multicolumn{4}{c}{\textit{Post-Hoc Debiasing (Frozen Backbone)}} \\
FairerCLIP~\cite{fairerclip} & ICLR'24    & named  & \xmark \\
RoboShot~\cite{roboshot}     & ICLR'24    & LLM    & \xmark \\
PRISM~\cite{prism}           & ICCV'25    & LLM    & \xmark \\
PRISM-mini~\cite{prism}      & ICCV'25    & LLM    & \xmark \\
$\text{SEM}_i$~\cite{sem}    & CVPR'26    & none   & \xmark \\
\rowcolor{oursrow}
\textbf{Ours}                & --         & none   & \cmark \\
\bottomrule
\end{tabular}
\caption{\textbf{Requirements across debiasing methods.} \textit{Bias Info} denotes the spurious attribute supervision required (per-sample labels, attributes \textbf{named} by the practitioner, \textbf{LLM} proposals, or \textbf{none}). \textit{Text-Agnostic} indicates methods that do not depend on a text encoder or prompt generation.}
\label{tab:taxonomy}
\end{table}
\section{Problem Setting}
\label{sec:problem}

Let each example in the dataset be a tuple $(x, y, a)$, where $x \in \mathcal{X}$ is an image, $y \in \mathcal{Y}$ is its class label, and $a \in \mathcal{A}$ is a latent bias attribute that is not causally related to $y$. Under a biased training distribution $P_{\text{train}}(x, y, a)$, $a$ and $y$ are spuriously correlated. A sample $x$ is \emph{bias-aligned} when the pair $(y, a)$ follows that spurious correlation and \emph{bias-conflicting} when it violates it. Let bias-ratio $\rho$ refer to the proportion of bias-conflicting samples in the dataset. A low $\rho$ denotes more biased dataset. 
A dataset may carry multiple bias attributes at once. We consider a setting where each class has at least one spurious attribute value that appears in the majority of its samples. The same attribute value may appear in other classes, but only in the minority of their samples. Multiple such attributes may exist for a single class. In this setting, two samples of \emph{different} classes that take the same value of $a$ are necessarily one aligned and one conflicting, since the majority value for one class can only be a minority value for the other. We call such a pair \textbf{bias-sharing}. Bias-sharing pairs are the ones a classifier is most likely to confuse. They agree on the bias attribute that the training correlation rewards and disagree on the class label.
We seek a classifier that performs well under a target distribution $P_{\text{test}}$ in which the spurious correlation is absent, and in particular perform well on the conflicting groups, where reliance on the bias attribute $a$ is penalized rather than being rewarded. The performance is therefore measured by worst-group accuracy alongside the overall accuracy.
In this work, we limit the problem setting to classifiers built on a frozen large pretrained vision encoder $f_I$ producing normalized embeddings $\mathbf{x} = f_I(x)$. The encoder is not fine-tuned or trained on the biased dataset. The objective is to learn a map applied to those embeddings, together with a linear head. 
We operate in the \emph{bias-unaware} setting, where the bias attribute $a$ is neither observed during training nor identified in advance. 
\begin{figure}[t]
\centering
\includegraphics[width=\linewidth]{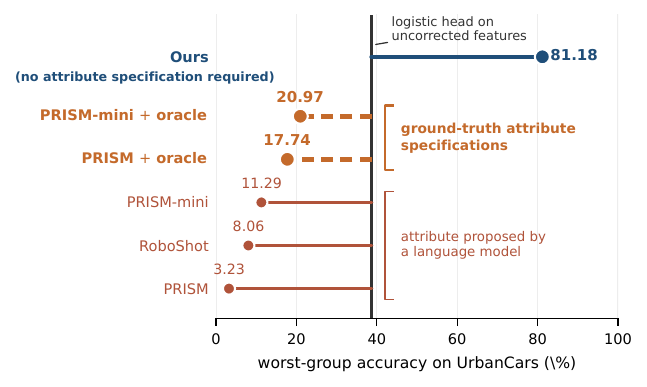}
\caption{\textbf{On UrbanCars, most debiasing methods are worse than no debiasing.} Worst-group accuracy of each method, measured against a linear classifier trained on the same frozen CLIP features with no debiasing. Bars to the left of the reference mark methods that reduce accuracy below it.
The dashed bar denotes PRISM with the ground-truth attribute identities
(\cref{subsec:oracle}).}
\label{fig:teaser}
\end{figure}
\begin{table}[t]
\centering
\small
\setlength{\tabcolsep}{3pt}
\renewcommand{\arraystretch}{1.15}
\begin{tabular}{l l cccc}
\toprule
\textbf{Method} & \textbf{Bias info}
& \textbf{Ovr.} & \textbf{W-BG} & \textbf{W-Co} & \textbf{W-Joint} \\
\midrule
\multicolumn{6}{l}{\emph{Cosine similarity to prompt embeddings}} \\
\rowcolor{refrow}
CLIP-cosine        & none    & 50.00 &  4.03 & 13.71 &  0.00 \\
PRISM              & LLM     & 53.43 &  6.45 & 16.13 &  1.61 \\
\rowcolor{oraclerow}
\quad $+$ \textbf{oracle}   & oracle  & 59.48 & 33.06 & 47.58 & 30.65 \\
PRISM-mini         & LLM     & 53.63 & 24.19 & 27.42 &  6.45 \\
\rowcolor{oraclerow}
\quad $+$ \textbf{oracle}   & oracle  & 55.65 & 26.61 & 20.97 &  9.68 \\
\midrule
\multicolumn{6}{l}{\emph{Linear head on the corrected features}} \\
\rowcolor{refrow}
CLIP-linear        & none    & 81.18 & 68.55 & 65.32 & 38.71 \\
PRISM              & LLM     & 61.49 & 32.30 & 33.10 &  3.23 \\
\rowcolor{oraclerow}
\quad $+$ \textbf{oracle}   & oracle  & 67.94 & 49.20 & 45.20 & 17.74 \\
PRISM-mini         & LLM     & 73.79 & 52.40 & 51.60 & 11.29 \\
\rowcolor{oraclerow}
\quad $+$ \textbf{oracle}   & oracle  & 76.61 & 56.45 & 57.26 & 20.97 \\
\bottomrule
\end{tabular}
\caption{\textbf{Correct specification is necessary but not sufficient.} Replacing LLM-generated attribute descriptions with ground-truth prompts improves PRISM and PRISM-mini, but both remain below the undebiased reference.}
\label{tab:oracle}
\end{table}

\begin{table}[t]
\centering
\small
\setlength{\tabcolsep}{6pt}
\renewcommand{\arraystretch}{1.15}
\begin{tabular}{l cc}
\toprule
\textbf{Cross-class pair shares\ldots}
& \textbf{all pairs} & \textbf{selected} \\
\midrule
neither shortcut           & 81.9 &  5.2 \\
background only            &  8.6 & 35.2 \\
co-occurring object only   &  8.6 & 12.1 \\
\textbf{both shortcuts}    & \textbf{0.9} & \textbf{47.6} \\
\midrule
at least one shortcut      & 18.1 & 94.8 \\
\midrule
number of pairs            & 16.0M & 986 \\
\bottomrule
\end{tabular}
\caption{\textbf{What a similarity threshold selects.} Composition of cross-class training pairs on UrbanCars (\%) in the frozen embedding space, that share each shortcut. \textit{Selected} denotes pairs whose cosine similarity exceeds the threshold used by our method (\cref{sec:method}). Attribute labels are used only for this analysis.}
\label{tab:composition}
\end{table}

\begin{figure}[t]
\centering
\includegraphics[width=0.9\linewidth]{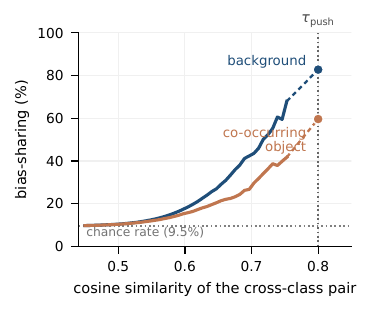}
\caption{ \textbf{Bias concentration scales with feature similarity.} The percentage of cross-class training pairs sharing a spurious background or co-occurring object increases monotonically as a function of the cosine similarity threshold.}
\label{fig:similarity}
\end{figure}
\section{Motivation}
\label{sec:analysis}

Post-hoc debiasing methods that rectify a frozen embedding space require the spurious attribute to be specified before they can act. Multiple such methods obtain this specification from a language model, which proposes candidates from class names alone. This section examines the impact of that requirement on a more realistic UrbanCars benchmark, where two spurious attributes are present at once. 
\subsection{Debiasing Performs Below the Baseline}
\label{subsec:belowref}



One might expect rectifying a biased representation to always help. Yet, we observe something strange on the UrbanCars benchmark (\cref{fig:teaser}). A linear classifier trained on features from a frozen pretrained model, with no debiasing, reaches $81.18\%$ overall and $38.71\%$ worst group accuracy. However, when we debias the features using post-hoc debiasing methods that use LLMs for bias-attribute specifications and train a same-capacity linear classifier on the debiased features, these methods perform very poorly. PRISM~\cite{prism} reaches $3.23\%$ worst-group accuracy, RoboShot~\cite{roboshot} reaches $8.06\%$, and PRISM-mini~\cite{prism} reaches $11.29\%$. Even the best method falls $27$ points below the naive classifier with no debiasing. Overall accuracy also declines, ruling out a robustness–accuracy trade-off.

\begin{figure*}[t]
\centering
\includegraphics[width=0.8\textwidth]{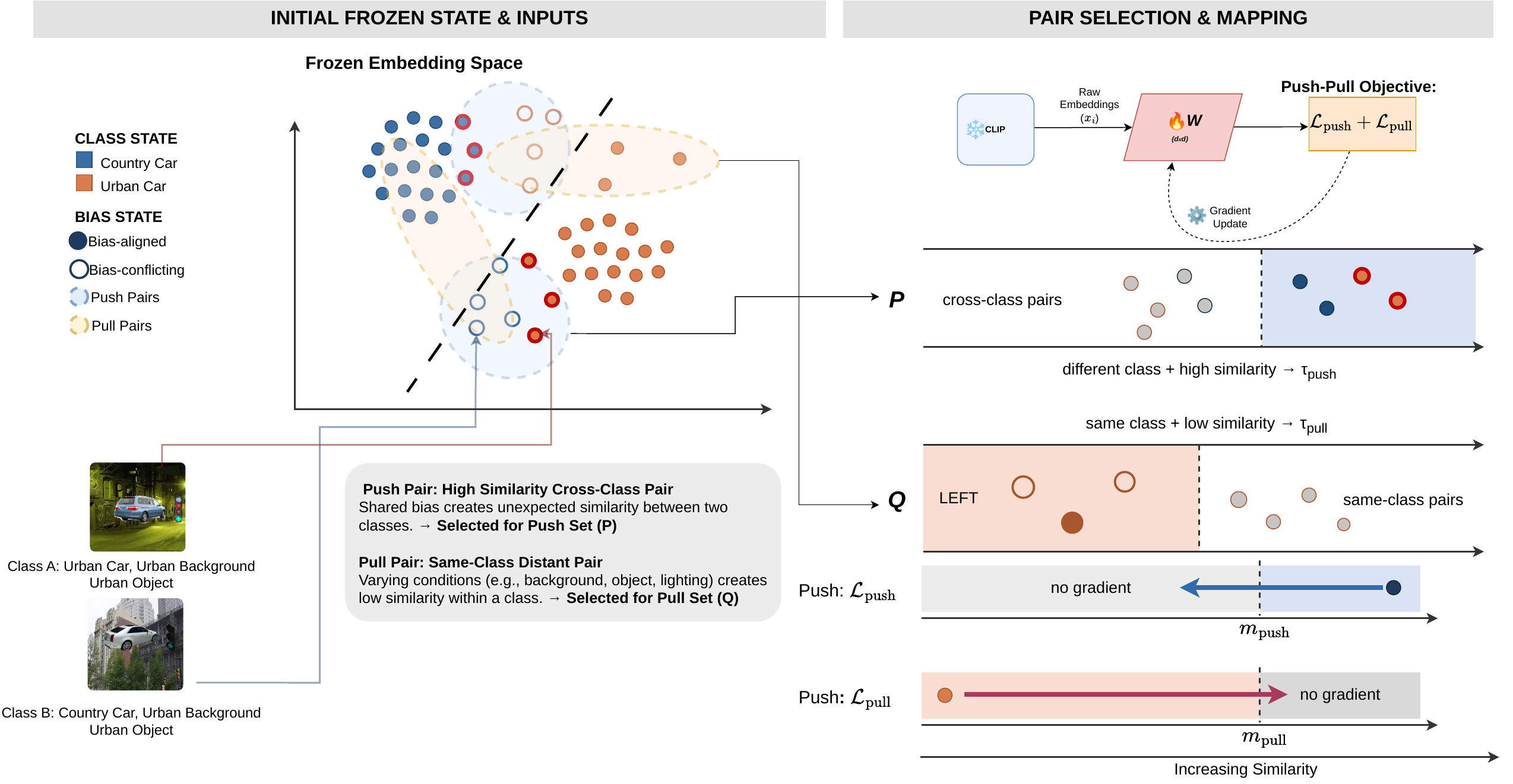}
\caption{\textbf{Method overview -- }\textbf{Left:} we select two sets of pairs in the frozen space using class labels and cosine similarity alone. $\mathcal{P}$ takes cross-class pairs above $\tau_{\mathrm{push}}$, $94.8\%$ of which are bias-sharing, and $\mathcal{Q}$ takes same-class pairs below $\tau_{\mathrm{pull}}$. \textbf{Right:} a selected pair contributes gradients until it reaches its margin and none after. Thresholds decide which pairs move, margins how far. We only train $\mathbf{W}$.}
\label{fig:method}
\end{figure*}
\subsection{Expert Bias Specification Is Not Enough}
\label{subsec:oracle}

Two competing explanations could account for the above failure. Firstly, LLM generated bias-specifications may not match the actual shortcuts. Secondly, the bias removal mechanism may be at fault. UrbanCars provides a direct test because it contains two co-occurring bias-shortcuts. We test the first explanation by replacing the LLM proposals with expert-written prompts that name the two ground-truth attributes, background and co-occurring object. Keeping the rest of the method same, we report the new results (+oracle) in \cref{tab:oracle}. Using a cosine classifier, PRISM improves from $1.61\%$ to $30.65\%$ on worst joint accuracy with the oracle information. With a trained linear head, the improvement is smaller, from $3.23\%$ to $17.74\%$. The performance never reaches the $38.71\%$ of the undebiased linear classifier. Since the ground truth specification helps in both cases, the specifications were indeed part of the problem. But the ground truth specification is insufficient to beat the baseline. A likely reason is that the two bias shortcuts do not occupy orthogonal subspaces in the frozen embedding space. If we represent each spurious attribute as the text derived direction of its prompt, excluding the class component they share by construction, the cosine similarity between these two directions is $0.58$. PRISM removes the direction associated with a specified attribute. When two directions overlap this much, removing one leaves part of the other behind, and the classifier shifts its reliance to that residual. Fixing one bias makes the other worse. This is the whac-a-mole failure described in \cite{urbancars}, and UrbanCars was built to expose this.

UrbanCars also exposes why the bias specification failed. Since an LLM proposes bias attributes for the given classes based on real-world co-occurrence priors, natural associations such as a vehicle type appearing against a specific background, get easily identified. However, the second shortcut in UrbanCars was introduced via image composition, corresponding to no predominant natural world observation. LLMs have nothing to draw upon in this regard, leaving them blind to artifacts that are not well-explored.


\subsection{Intrinsic Trace of Bias in the Feature Space Geometry}
\label{subsec:geometry}
Prompt-based methods require bias specification because they treat the spurious attribute as an external semantic concept that must be described before it can be eliminated. However, the bias attribute also leaves an intrinsic geometric signature in the representation that can be precisely measured without requiring a semantic description
\paragraph{Spurious features perturb the embedding space almost as much as class boundaries.}
In the frozen embedding space, two images in UrbanCars of the same class on the same background have a mean cosine similarity of $0.630$. We analyze the effect of changing one attribute at a time. Changing only the background reduces this similarity to $0.589$. Changing only the class while keeping the background same reduces it to $0.584$. Therefore, a change in spurious attribute reduces cosine similarity by almost the same amount as changing the class. Therefore, a cross-class \textbf{bias-sharing} pair of samples that have the same background, will be placed almost as close to each other as two same class samples, by the frozen encoder. We study this finding to discover a geometric signature of the bias.
\paragraph{High-similarity cross-class pairs are overwhelmingly bias-sharing.}
The decision boundary depends on the pairs of samples the encoder places closest together, not on the average pair. We refer to the two spurious attributes in UrbanCars, background and co-occurring object, as shortcuts. A cross-class pair drawn at random shares neither shortcut in $81.9\%$ of cases, shares one in $17.2\%$ cases, and both in only $0.9\%$ cases. When we select only the pairs with high cosine similarity, this distribution reverses (\cref{tab:composition}). Pairs sharing neither shortcut fall to $5.2\%$, while pairs sharing both rise to $47.6\%$, and pairs sharing atleast one shortcut rises to 94.8\%, becoming the dominant configuration. Therefore, the pairs the encoder finds most confusable due to high similarity are almost always \textbf{bias-sharing}. The reversal strengthens as the similarity threshold increases (\cref{fig:similarity}). The proportion of selected pairs that share the background or the co-occurring object rises continuously with the cutoff. This is an intrinsic property of the decision boundary, not an artifact of a chosen threshold. Selection therefore requires no prior knowledge of which attribute makes a pair similar or of how many attributes are involved. It is a measurable property of the representation, and we can use this to learn a transformation that separates bias-sharing pairs to debias the features.



\section{Proposed Approach}
\label{sec:method}

The analysis in \Cref{subsec:geometry} revealed an intrinsic geometric property: the cross-class pairs that a frozen encoder finds most confusable (high cosine similarity) are almost all \textbf{bias-sharing}. We turn this observation into an objective for post-hoc debiasing. The proposed approach learns a linear map $\mathbf{W} \in \mathbb{R}^{d \times d}$ which transforms the features of the frozen pretrained encoder to debias them (\cref{fig:method}).

\subsection{Bias-sharing pair selection}
\label{subsec:selection}
 
Let $\mathbf{x}_i$ denote the frozen embedding of image $i$ produced by the pretrained encoder, and let $y_i$ denote its class label. We consider pairs of training images $(i,j)$ and measure their cosine similarity as $\langle \mathbf{x}_i, \mathbf{x}_j \rangle$. Motivated by our findings in \cref{subsec:geometry}, we present the following condition for selecting bias-sharing pairs from the cross-class pairs. 
\begin{equation}
  \label{eq:push_set}
  \mathcal{P} = \big\{(i,j) : y_i \neq y_j,\;
      \langle \mathbf{x}_i, \mathbf{x}_j \rangle > \tau_{\text{push}}\big\}.
\end{equation}
As characterized in \cref{tab:composition}, applying this threshold on the UrbanCars dataset yields a set $\mathcal{P}$ where $94.8\%$ of the members are bias-sharing, compared to just $18.1\%$ in the unconstrained cross-class distribution. Because every pair in $\mathcal{P}$ spans two distinct classes, an objective built solely on this set would leave the geometry \emph{within} each individual class unconstrained. To anchor the intra-class representation, we sample a secondary set of distant same-class pairs:
\begin{equation}
  \label{eq:pull_set}
  \mathcal{Q} \subseteq \big\{(i,j) : y_i = y_j,\;
      \langle \mathbf{x}_i, \mathbf{x}_j \rangle < \tau_{\text{pull}}\big\},
\end{equation}
taking one partner per image and retaining $|\mathcal{P}|/N$ pairs per class (N refers to the number of classes), so that the two sets contribute comparable numbers of pairs for balanced regularization. Since bias-conflicting samples constitute a severe minority of the data, the overwhelming majority of any given class is bias-aligned.
Consequently, the low-similarity regime of $\mathcal{Q}$ predominantly captures natural intra-class variance among these abundant aligned samples. However, this set also captures the rare/minor instances where the spurious attribute itself differs (pairing an aligned sample with a conflicting one). Pulling these pairs together is highly beneficial, as it encourages invariance to the spurious attribute while preserving overall class cohesion. Therefore, while $\mathcal{P}$ can be processed to actively separate the confusable cross-class pairs, $\mathcal{Q}$ acts as a broad structural anchor that prevents the internal geometry of the classes from collapsing. We discuss the contribution of each set in (\cref{subsec:ablations}). Both sets are computed once, from the frozen embeddings, before $\mathbf{W}$ exists. Membership is fixed by the geometry measured in \cref{sec:analysis} rather than by the representation the method is producing.
 

\subsection{Managing bias-sharing pairs and low similarity same class pairs}
\label{subsec:objective}

 Let $\mathbf{z}_i = \mathbf{W}\mathbf{x}_i / \lVert \mathbf{W}\mathbf{x}_i \rVert$ denote the projected embedding and $s_{ij} = \langle \mathbf{z}_i, \mathbf{z}_j \rangle$ the similarity it induces. The sets $\mathcal{P}$ and $\mathcal{Q}$ are used to optimize $\mathbf{W}$ with a one-sided hinge against a margin:
\begin{align}
  \label{eq:push_loss}
  \mathcal{L}_{\text{push}} &= \frac{1}{|\mathcal{P}|}
    \sum_{(i,j)\in\mathcal{P}}
    \big[\, s_{ij} - m_{\text{push}} \,\big]_+ , \\
  \label{eq:pull_loss}
  \mathcal{L}_{\text{pull}} &= \frac{1}{|\mathcal{Q}|}
    \sum_{(i,j)\in\mathcal{Q}}
    \big[\, m_{\text{pull}} - s_{ij} \,\big]_+ , \\
  \label{eq:total_loss}
  \mathcal{L} &= \mathcal{L}_{\text{push}} + \mathcal{L}_{\text{pull}} ,
\end{align}
with $m_{\text{push}} < \tau_{\text{push}}$ and $m_{\text{pull}} > \tau_{\text{pull}}$.

A pair contributes gradient to optimize $\mathbf{W}$ until its similarity reaches the margin and none thereafter. The thresholds determine which pairs are moved and the margins determine how far they are moved. Replacing both hinges with unbounded terms costs $4.84$ points of Worst Joint accuracy on UrbanCars (see \cref{tab:ablation}).

Note that \crefrange{eq:push_set}{eq:total_loss} do not use any information regarding the bias attribute, and no text embedding or auxiliary models are used for our proposed approach. The proposed objective only utilizes image-to-image similarity and class labels alone to learn a linear map to debias the features of a frozen large pretrained model.

\subsection{Overview and Inference}
\label{subsec:optimization}


In the proposed approach, we initialize the linear map $\mathbf{W} \in \mathbb{R}^{d\times d}$ as an identity matrix, ensuring optimization begins from the encoder's native geometry. It is the sole component that receives a gradient, introducing only $d^2 \approx 0.59$M learnable parameters compared to the backbone's ${\sim}300$M. After learning the projection, we fit a linear classifier on the training split while keeping $\mathbf{W}$ fixed. Since the CLIP+linear-head baseline in \cref{subsec:main_results} employs an identical head and training schedule on unprojected embeddings, the two setups differ exclusively by the presence of $\mathbf{W}$. The objective $\mathcal{L}$ depends on $\mathbf{W}$ only through cosine similarities between transformed embeddings, so $\mathbf{W}$'s scale is irrelevant, and the optimization effectively learns a linear map that reshapes pairwise similarities in the frozen feature space. Performance gains come from the geometric objective guiding $\mathbf{W}$ toward a more robust representation.

\begin{table}[t]
\centering
\small
\setlength{\tabcolsep}{1pt}
\renewcommand{\arraystretch}{1.15}
\begin{tabular}{l l cccc}
\toprule
\textbf{Method} & \textbf{Venue}
& \textbf{Ovr.} & \textbf{W-BG} & \textbf{W-Co} & \textbf{W-Joint} \\
\midrule
\multicolumn{6}{l}{\emph{Cosine similarity to prompt embeddings}} \\
\rowcolor{refrow}
CLIP-cos        & --      & 50.00 &  4.03 & 13.71 &  0.00 \\
RoboShot           & ICLR'24 & 51.81 & 13.71 & 20.16 &  0.00 \\
FairerCLIP         & ICLR'24 & \second{81.25} & 64.52 & 69.36 & 58.06 \\
PRISM              & ICCV'25 & 53.43 &  6.45 & 16.13 &  1.61 \\
PRISM-m        & ICCV'25 & 53.63 & 24.19 & 27.42 &  6.45 \\
$^{\dagger}$$\text{SEM}_i$                & CVPR'26 & 78.63 & \second{69.35} & \second{71.77} & \second{61.29} \\
\rowcolor{oursrow}
\textbf{Ours}      & --      & \best{89.78}\std{0.41} & \best{85.48}\std{1.74} & \best{84.95}\std{1.66} & \best{81.18}\std{2.28} \\
\midrule
\multicolumn{6}{l}{\emph{Linear head on the corrected features}} \\
\rowcolor{refrow}
CLIP-L       & --      & 81.18 & 68.55 & 65.32 & 38.71 \\
RoboShot           & ICLR'24 & 72.40 & 48.39 & \second{51.61} &  8.06 \\
FairerCLIP         & ICLR'24 & 50.81 &  9.68 & 41.94 &  8.06 \\
PRISM              & ICCV'25 & 61.49 & 32.30 & 33.10 &  3.23 \\
PRISM-m         & ICCV'25 & \second{73.79} & \second{52.40} & 51.60 & \second{11.29} \\
\rowcolor{oursrow}
\textbf{Ours}      & --      & \best{90.59}\std{0.71} & \best{85.22}\std{1.23} & \best{83.60}\std{0.93} & \best{81.18}\std{0.93} \\
\bottomrule
\end{tabular}
\caption{\textbf{UrbanCars: two simultaneous shortcuts.}
Methods are grouped by the rule used to classify, so that comparison is made between methods sharing a decision procedure. Shaded grey rows are the undebiased reference for each rule. Best method: \best{bold}, second: \second{underlined}. Results are averaged over 3 seeds. For our method, we report the mean and standard deviation.$^{\dagger}$$\text{SEM}_i$ transforms only the text embeddings leaving image features unchanged, so a head fit on its features reproduces CLIP-linear exactly.}
\label{tab:urbancars}
\end{table}
\begin{table*}[t]
\centering
\small
\setlength{\tabcolsep}{2.5pt}
\renewcommand{\arraystretch}{1.15}
\resizebox{\textwidth}{!}{
\begin{tabular}{l l l cc cc @{\hskip 6pt} cc cc cc cc @{\hskip 6pt} cc}
\toprule
& & & \multicolumn{4}{c}{\textbf{BAR}} & \multicolumn{8}{c}{\textbf{BFFHQ}}
  & \multicolumn{2}{c}{\textbf{Waterbirds}$^{\ddagger}$} \\
\cmidrule(lr){4-7}\cmidrule(lr){8-15}\cmidrule(lr){16-17}
& & & \multicolumn{2}{c}{$1\%$} & \multicolumn{2}{c}{$5\%$}
  & \multicolumn{2}{c}{$0.5\%$} & \multicolumn{2}{c}{$1\%$}
  & \multicolumn{2}{c}{$2\%$} & \multicolumn{2}{c}{$5\%$} & \multicolumn{2}{c}{$5\%$} \\
\cmidrule(lr){4-5}\cmidrule(lr){6-7}\cmidrule(lr){8-9}\cmidrule(lr){10-11}\cmidrule(lr){12-13}\cmidrule(lr){14-15}\cmidrule(lr){16-17}
\textbf{Method} & \textbf{Venue} & \textbf{Bias info}
& Ovr. & Worst & Ovr. & Worst & Ovr. & Worst & Ovr. & Worst
& Ovr. & Worst & Ovr. & Worst & Ovr. & Worst \\
\midrule
\multicolumn{17}{l}{\emph{Cosine similarity to prompt embeddings}} \\
\rowcolor{refrow}
CLIP-cosine   & --      & none  & 96.49 & 74.65 & 96.49 & 74.65 & 70.70 & 37.60 & 70.70 & 37.60 & 70.70 & 37.60 & 70.70 & 37.60 & 86.20 & 39.60 \\
RoboShot\cite{roboshot}      & ICLR'24 & LLM   & 24.08 &  0.00 & 24.08 &  0.00 & 47.90 &  2.80 & 47.90 &  2.80 & 47.90 &  2.80 & 47.90 &  2.80 & 72.10 & 41.90 \\
FairerCLIP\cite{fairerclip}    & ICLR'24 & named & 94.03 & 59.16 & 82.95 & 21.13 & 73.50 & 38.00 & 83.40 & 56.80 & \second{94.00} & 83.60 & \second{94.40} & 83.60 & 89.12 & \second{77.41} \\
PRISM\cite{prism}         & ICCV'25 & LLM   & 95.96 & \second{76.06} & 95.96 & \second{76.06} & 91.80 & \second{87.60} & 91.80 & \second{87.60} & 91.80 & 87.60 & 91.80 & 87.60 & \second{89.18} & {67.91} \\
PRISM-mini\cite{prism}    & ICCV'25 & LLM   & \second{96.49} & 74.65 & \second{96.49} & 74.65 & 73.60 & 39.20 & 73.60 & 39.20 & 73.60 & 39.20 & 73.60 & 39.20 & 87.23 & 53.43 \\
$\text{SEM}_i$\cite{sem} & CVPR'26 & none  & 95.08 & 66.20 & 95.08 & 66.20 & \second{92.00} & \best{88.40} & \second{92.00} & \second{88.40} & 92.00 & \second{88.40} & 92.00 & \second{88.40} & 83.20 & 52.30 \\
\rowcolor{oursrow}
\textbf{Ours} & --      & none  & \best{97.31} & \best{87.79} & \best{97.83} & \best{87.79} & \best{95.03} & \second{88.00} & \best{95.50} & \best{90.80} & \best{96.60} & \best{92.67} & \best{97.70} & \best{95.47} & \best{89.29} & \best{81.95} \\
\midrule
\multicolumn{17}{l}{\emph{Linear head on the corrected features}} \\
\rowcolor{refrow}
CLIP-linear   & --      & none  & 96.13 & 70.42 & 97.19 & 78.87 & 90.80 & 78.00 & 92.00 & 82.80 & 94.30 & 87.60 & 96.40 & 92.00 & 88.47 & \second{73.67} \\
RoboShot\cite{roboshot}      & ICLR'24 & LLM   & 92.10 & 54.90 & 93.70 & 63.40 & 79.10 & 56.00 & 84.50 & 67.60 & 91.60 & 78.80 & 95.70 & 88.40 & 83.50 & 63.60 \\
FairerCLIP\cite{fairerclip}    & ICLR'24 & named & 71.35 & 16.81 & 74.34 & 59.66 & 48.20 & 42.40 & 83.70 & 53.20 & 93.60 & 88.80 & 96.50 & 92.00 & \best{89.71} & 60.90 \\
PRISM\cite{prism}         & ICCV'25 & LLM   & \second{97.19} & \second{81.69} & \second{97.19} & \second{83.10} & \second{91.80} & \second{83.20} & \second{93.90} & \second{84.80} & \second{95.10} & \second{89.60} & \second{96.60} & \second{92.80} & 82.95 & 64.79 \\
PRISM-mini\cite{prism}    & ICCV'25 & LLM   & 94.02 & 57.75 & 95.08 & 61.97 & 80.70 & 57.20 & 86.50 & 72.80 & 92.00 & 79.60 & 96.00 & 88.80 & 85.42 & 70.64 \\
\rowcolor{oursrow}
\textbf{Ours} & --      & none  & \best{97.54} & \best{83.10} & \best{97.89} & \best{84.51} & \best{92.67} & \best{84.00} & \best{94.30} & \best{86.67} & \best{95.60} & \best{90.67} & \best{97.93} & \best{95.87} & \second{89.35} & \best{80.01} \\
\bottomrule
\end{tabular}
}
\caption{Results on the BAR, BFFHQ, and Waterbirds datasets. For our method, we report the mean and standard deviation. Note: \textsuperscript{\textdaggerdbl} For Waterbirds, we follow the evaluation protocol of \cite{sem}, including the ViT-L/14@336px backbone. Further details, including standard deviations, are provided in the supplementary materials.}
\label{tab:bar_bffhq}
\end{table*}

\section{Experiments}
\label{sec:experiments}

\subsection{Setup}
\label{subsec:exp_setup}

\paragraph{Datasets.}
We evaluate on four benchmarks. \textbf{UrbanCars}~\cite{urbancars} is a binary vehicle classification task constructed with \emph{two} simultaneous shortcuts, image background (BG), and a co-occurring object (CoObj). \textbf{BAR}~\cite{lff} is a six-class action recognition task in which each action co-occurs with a typical place, evaluated at bias ratios $\rho= $1\% and 5\%. \textbf{BFFHQ}~\cite{bffhq1,bffhq2} is an age classification task with a demographic shortcut, evaluated at $\rho=$0.5\%, 1\%, 2\%, and 5\%. \textbf{Waterbirds}~\cite{groupdro} is a bird classification task with a background shortcut, evaluated at $\rho=$5\%.

\paragraph{Metrics and Protocol.}
\label{para:metrics}
We report overall and worst-group accuracy. On UrbanCars we separate worst-group accuracy (WGA) by shift type: \emph{Worst BG} (W-BG) and \emph{Worst CoObj} (W-Co) are the lowest accuracy over groups in which the respective attribute is shifted, and \emph{Worst Joint} (W-Joint) over groups in which both are shifted at once. Worst Joint is the quantity that multiple shortcuts affect directly. Hyperparameters are selected per dataset on a held-out validation split, and test results are reported for the selected configuration. We run each experiment with three random seeds and report the mean and standard deviation. Further implementation details are provided in the supplementary materials.

\paragraph{Baselines.}
All methods use the same frozen CLIP ViT-L/14 backbone. We include two undebiased references: CLIP with cosine similarity to prompt embeddings (\emph{CLIP-cosine}), and a linear classifier trained on the frozen embeddings (\emph{CLIP-linear}). CLIP-linear is the matched-capacity reference for our method, using an identical head, optimizer, and scheduler, and differing only in the absence of $\mathbf{W}$. We report our method under both classification rules. The \emph{cosine} variant applies $\mathbf{W}$ to image embeddings and performs classification by cosine similarity with the same frozen prompt embeddings used by the vanilla-cosine classifier.
We compare against $\text{SEM}_i$~\cite{sem}, PRISM and PRISM-mini~\cite{prism}, RoboShot~\cite{roboshot} and FairerCLIP~\cite{fairerclip}. PRISM and RoboShot obtain their attributes from an LLM without human specification. 
For every competing method, we tune all of its hyperparameters and report the best configuration found. No method receives per-sample attribute annotations.

\subsection{Main results}
\label{subsec:main_results}


\paragraph{Multiple shortcuts.}
On the UrbanCars benchmark, every method that obtains its attribute from an LLM falls below the undebiased linear classifier on all four metrics (\cref{tab:urbancars}). Our method achieves $81.18\%$ Worst Joint accuracy, outperforming the best competing method by $19.89$ points and CLIP-linear by $42.47$ points. The gap between overall and Worst Joint accuracy shrinks from $45.16$ points for the baseline to $9.41$ points for our method, while overall accuracy improves from $81.18\%$ to $90.59\%$. Unlike text-based baselines, our method also applies to vision-only backbones such as DINOv3. We report these results in the supplementary materials.


\paragraph{Single shortcuts.}
The advantage extends to single-shortcut benchmarks (\cref{tab:bar_bffhq}). On BAR, our method achieves the highest worst-group accuracy under both classification rules at both bias ratios. Under the linear head, we reach $83.10\%$ at $1\%$ and $84.51\%$ at $5\%$, exceeding the best competitor by $1.41$ points at each ratio. Under the cosine rule, the margin widens. We reach $87.79\%$ at $1\%$, exceeding PRISM by $11.73$ points. On BFFHQ, our WGA scales with the availability of conflicting samples, rising from $84.00\%$ at $0.5\%$ to $95.87\%$ at $5\%$ under the linear head. PRISM improves from $83.20\%$ to $92.80\%$ over the same range, but our method gains $11.87$ points while PRISM gains $9.60$ points. Our pair-selection strategy queries the dataset's own geometry, so it naturally improves as more conflicting samples become available. On Waterbirds, our method reaches $80.01\%$ WGA under the linear head, surpassing PRISM by $15.22$ points and FairerCLIP by $19.11$ points. Under the cosine rule, we reach $81.95\%$, exceeding FairerCLIP by $4.54$ points and PRISM by $14.04$ points.
\begin{table}[t]
\centering
\small
\setlength{\tabcolsep}{2pt}
\renewcommand{\arraystretch}{1.15}
\begin{tabular}{ccc cccc}
\toprule
\multicolumn{3}{c}{\textbf{Objective}} & \multicolumn{4}{c}{\textbf{UrbanCars}} \\
\cmidrule(lr){1-3}\cmidrule(lr){4-7}
$\mathcal{L}_{\text{push}}$ & $\mathcal{L}_{\text{pull}}$ & margins
& Ovr. & W-BG & W-Co & W-Joint \\
\midrule
\cmark & --     & \cmark & 85.48 & 73.39 & 70.97 & 48.39 \\
--     & \cmark & \cmark & 71.77 & 47.58 & 49.19 &  6.45 \\
\cmark & --     & --     & 87.10 & 78.23 & 75.81 & 58.06 \\
--     & \cmark & --     & 87.70 & 78.23 & 75.81 & 61.29 \\
\cmark & \cmark & --     & 90.73 & 86.29 & 84.68 & 79.03 \\
\rowcolor{oursrow}
\cmark & \cmark & \cmark & \best{90.93} & \best{86.29} & \best{85.48} & \best{83.87} \\
\bottomrule
\end{tabular}
\caption{\textbf{Ablation on loss terms}}
\label{tab:ablation}
\end{table}

\begin{figure}[t]
\centering
\includegraphics[width=0.9\linewidth]{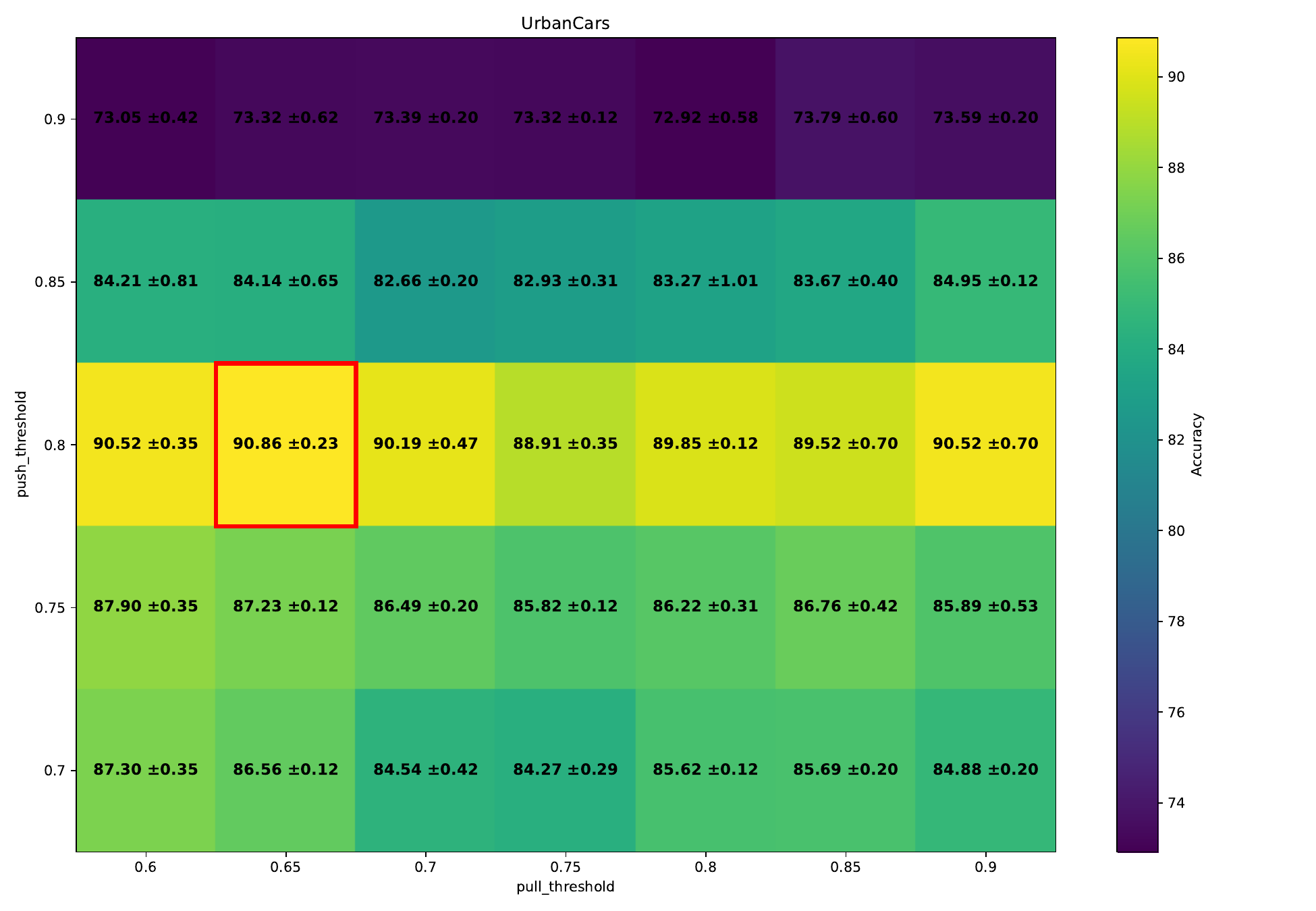}
\caption{\textbf{Threshold stability on UrbanCars.} Each cell reports the mean and spread across the four margin settings at that threshold pair. The red border box marks the selected configuration.}
\label{fig:sensitivity}
\end{figure}




\subsection{Ablations}
\label{subsec:ablations}

Our objective combines two forces. The first separates bias-sharing cross-class pairs ($\mathcal{L}_{\text{push}}$), and the second maintains same-class cohesion ($\mathcal{L}_{\text{pull}}$). \cref{tab:ablation} shows that both are necessary. Removing either term drops Worst Joint accuracy by at least $22.58$ points. The margins prevent unbounded optimization and add a further $4.84$ points on top of the full objective. \cref{fig:sensitivity} Sweeps the two thresholds on UrbanCars. We use four random margin values for each threshold pair and average the results, reporting the mean and standard deviation. Corresponding grids for BAR, BFFHQ, and Waterbirds are provided in the supplementary materials. Additional analyses on backbones, class imbalance, and the separation of same-class and cross-class pairs after projection are in the supplementary materials.

\section{Conclusion}
We have shown that debiasing a frozen encoder need not require explicit spurious-attribute identification. The bias signal is already encoded in the frozen feature geometry. Highly similar cross-class pairs are overwhelmingly bias-sharing, and this is detectable from pairwise similarity and class labels alone. Our method turns this observation into a training objective that separates bias-sharing pairs and maintains same-class cohesion. It consistently outperforms post-hoc debiasing approaches, including on the multi-bias benchmark, where methods dependent on LLM-generated specifications underperform an undebiased linear classifier.

{
    \small
    \bibliographystyle{ieeenat_fullname}
    \bibliography{main}

@String(CVPR= {IEEE Conf. Comput. Vis. Pattern Recog.})

@String(ICCV= {Int. Conf. Comput. Vis.})

@String(ICLR = {Int. Conf. Learn. Represent.})

@String(CVPR  = {CVPR})

@String(ICCV  = {ICCV})

@String(ICLR  = {ICLR})

@inproceedings{
groupdro,
title={Distributionally Robust Neural Networks},
author={Shiori Sagawa* and Pang Wei Koh* and Tatsunori B. Hashimoto and Percy Liang},
booktitle={International Conference on Learning Representations},
year={2020},
url={https://openreview.net/forum?id=ryxGuJrFvS}
}

@inproceedings{
dfr,
title={Towards Last-Layer Retraining for Group Robustness with Fewer Annotations},
author={Tyler LaBonte and Vidya Muthukumar and Abhishek Kumar},
booktitle={Thirty-seventh Conference on Neural Information Processing Systems},
year={2023},
url={https://openreview.net/forum?id=kshC3NOP6h}
}

@inproceedings{lff,
 author = {Nam, Junhyun and Cha, Hyuntak and Ahn, Sungsoo and Lee, Jaeho and Shin, Jinwoo},
 booktitle = {Advances in Neural Information Processing Systems},
 editor = {H. Larochelle and M. Ranzato and R. Hadsell and M.F. Balcan and H. Lin},
 pages = {20673--20684},
 publisher = {Curran Associates, Inc.},
 title = {Learning from Failure: De-biasing Classifier from Biased Classifier},
 url = {https://proceedings.neurips.cc/paper_files/paper/2020/file/eddc3427c5d77843c2253f1e799fe933-Paper.pdf},
 volume = {33},
 year = {2020}
}

@InProceedings{jtt,
  title = 	 {Just Train Twice: Improving Group Robustness without Training Group Information},
  author =       {Liu, Evan Z and Haghgoo, Behzad and Chen, Annie S and Raghunathan, Aditi and Koh, Pang Wei and Sagawa, Shiori and Liang, Percy and Finn, Chelsea},
  booktitle = 	 {Proceedings of the 38th International Conference on Machine Learning},
  pages = 	 {6781--6792},
  year = 	 {2021},
  editor = 	 {Meila, Marina and Zhang, Tong},
  volume = 	 {139},
  series = 	 {Proceedings of Machine Learning Research},
  month = 	 {18--24 Jul},
  publisher =    {PMLR},
  url = 	 {https://proceedings.mlr.press/v139/liu21f.html}
}

@InProceedings{urbancars,
    author    = {Li, Zhiheng and Evtimov, Ivan and Gordo, Albert and Hazirbas, Caner and Hassner, Tal and Ferrer, Cristian Canton and Xu, Chenliang and Ibrahim, Mark},
    title     = {A Whac-a-Mole Dilemma: Shortcuts Come in Multiples Where Mitigating One Amplifies Others},
    booktitle = {Proceedings of the IEEE/CVF Conference on Computer Vision and Pattern Recognition (CVPR)},
    month     = {June},
    year      = {2023},
    pages     = {20071-20082}
}

@inproceedings{prism,
  title={Prism: Reducing spurious implicit biases in vision-language models with llm-guided embedding projection},
  author={Molahasani, Mahdiyar and Motamedi, Azadeh and Greenspan, Michael and Kim, Il-Min and Etemad, Ali},
  booktitle={Proceedings of the IEEE/CVF International Conference on Computer Vision},
  pages={688--697},
  year={2025}
}

@inproceedings{fairerclip,
  title={FairerCLIP: Debiasing CLIP's Zero-Shot Predictions using Functions in RKHSs},
  author={Dehdashtian, Sepehr and Wang, Lan and Boddeti, Vishnu Naresh},
  booktitle={International Conference on Learning Representations (ICLR)},
  year={2024}
}

@inproceedings{sem,
  title={{SEM: Sparse Embedding Modulation for Post-Hoc Debiasing of Vision-Language Models}},
  author={Guimard, Quentin and Bartsch, Federico and Caldarella, Simone and Aljundi, Rahaf and Ricci, Elisa and Mancini, Massimiliano},
  booktitle={Findings of the IEEE/CVF Conference on Computer Vision and Pattern Recognition (CVPR)},
  year={2026}
}

@inproceedings{
roboshot,
title={Zero-Shot Robustification of Zero-Shot Models},
author={Dyah Adila and Changho Shin and Linrong Cai and Frederic Sala},
booktitle={The Twelfth International Conference on Learning Representations},
year={2024},
url={https://openreview.net/forum?id=fCeUoDr9Tq}
}

@inproceedings{clip,
  title={Learning transferable visual models from natural language supervision},
  author={Radford, Alec and Kim, Jong Wook and Hallacy, Chris and Ramesh, Aditya and Goh, Gabriel and Agarwal, Sandhini and Sastry, Girish and Askell, Amanda and Mishkin, Pamela and Clark, Jack and others},
  booktitle={International conference on machine learning},
  pages={8748--8763},
  year={2021},
  organization={PMLR}
}

@ARTICLE{bffhq1,
  author={Karras, Tero and Laine, Samuli and Aila, Timo},
  journal={IEEE Transactions on Pattern Analysis and Machine Intelligence}, 
  title={A Style-Based Generator Architecture for Generative Adversarial Networks}, 
  year={2021},
 }

@INPROCEEDINGS{bffhq2,
  author={Kim, Eungyeup and Lee, Jihyeon and Choo, Jaegul},
  booktitle={2021 IEEE/CVF International Conference on Computer Vision (ICCV)}, 
  title={BiaSwap: Removing Dataset Bias with Bias-Tailored Swapping Augmentation}, 
  year={2021}}

@InProceedings{siglip,
    author    = {Zhai, Xiaohua and Mustafa, Basil and Kolesnikov, Alexander and Beyer, Lucas},
    title     = {Sigmoid Loss for Language Image Pre-Training},
    booktitle = {Proceedings of the IEEE/CVF International Conference on Computer Vision (ICCV)},
    month     = {October},
    year      = {2023},
    pages     = {11975-11986}
}

@article{
dinov3,
title={{DINO}v3},
author={Oriane Sim{\'e}oni and Huy V. Vo and Maximilian Seitzer and Federico Baldassarre and Maxime Oquab and Cijo Jose and Vasil Khalidov and Marc Szafraniec and Seung Eun Yi and Michael Ramamonjisoa and Francisco Massa and Daniel HAZIZA and Luca Wehrstedt and Jianyuan Wang and Timoth{\'e}e Darcet and Th{\'e}o Moutakanni and Leonel Sentana and Claire Roberts and Andrea Vedaldi and Jamie Tolan and John Brandt and Camille Couprie and Julien Mairal and Herve Jegou and Patrick Labatut and Piotr Bojanowski},
journal={Transactions on Machine Learning Research},
issn={2835-8856},
year={2026},
}

@InProceedings{bias1,
    author    = {Meister, Nicole and Zhao, Dora and Wang, Angelina and Ramaswamy, Vikram V. and Fong, Ruth and Russakovsky, Olga},
    title     = {Gender Artifacts in Visual Datasets},
    booktitle = {Proceedings of the IEEE/CVF International Conference on Computer Vision (ICCV)},
    month     = {October},
    year      = {2023},
    pages     = {20267-20277}
}

@inproceedings{bias2,
    title = "Are Gender-Neutral Queries Really Gender-Neutral? Mitigating Gender Bias in Image Search",
    author = "Wang, Jialu  and
      Liu, Yang  and
      Wang, Xin",
    booktitle = "Proceedings of the 2021 Conference on Empirical Methods in Natural Language Processing",
    month = nov,
    year = "2021",
    publisher = "Association for Computational Linguistics",
}

@inproceedings{bias3,
author = {Yang, Kaiyu and Qinami, Klint and Fei-Fei, Li and Deng, Jia and Russakovsky, Olga},
title = {Towards fairer datasets: filtering and balancing the distribution of the people subtree in the ImageNet hierarchy},
year = {2020},
publisher = {Association for Computing Machinery},
booktitle = {Proceedings of the 2020 Conference on Fairness, Accountability, and Transparency},
pages = {547–558},
numpages = {12},
location = {Barcelona, Spain},
series = {FAT* '20}
}

@inproceedings{bias4,
    title = "{VLS}tereo{S}et: A Study of Stereotypical Bias in Pre-trained Vision-Language Models",
    author = "Zhou, Kankan  and
      Lai, Eason  and
      Jiang, Jing",
    booktitle = "Proceedings of the 2nd Conference of the Asia-Pacific Chapter of the Association for Computational Linguistics and the 12th International Joint Conference on Natural Language Processing (Volume 1: Long Papers)",
    month = nov,
    year = "2022",
    address = "Online only",
    publisher = "Association for Computational Linguistics",
    url = "https://aclanthology.org/2022.aacl-main.40/",
    doi = "10.18653/v1/2022.aacl-main.40",
    pages = "527--538",
}

@misc{bias5,
      title={Understanding Bias in Large-Scale Visual Datasets}, 
      author={Boya Zeng and Yida Yin and Zhuang Liu},
      year={2024},
      eprint={2412.01876},
      archivePrefix={arXiv},
      primaryClass={cs.CV},
      url={https://arxiv.org/abs/2412.01876}, 
}

@article{bias6,
  author    = {Wang, Angelina and Liu, Alexander and Zhang, Ryan and Kleiman, Anat and Kim, Leslie and Zhao, Dora and Shirai, Iroha and Narayanan, Arvind and Russakovsky, Olga},
  title     = {REVISE: A Tool for Measuring and Mitigating Bias in Visual Datasets},
  journal   = {International Journal of Computer Vision},
  volume    = {130},
  number    = {7},
  pages     = {1790--1810},
  year      = {2022},
  publisher = {Springer},
  doi       = {10.1007/s11263-022-01625-5}
}
}

\end{document}


\maketitle
\section{Implementation Details}
\label{sec:supp_impl}

\Cref{alg:method_supp} states the procedure in full.
Pair selection runs once, over the cached embeddings of the frozen encoder, and is the only stage whose cost grows with the size of the training split. Learning $\mathbf{W}$ then operates on the selected pairs alone.

\paragraph{Training.}
$\mathbf{W}\in\mathbb{R}^{d\times d}$ is initialized at $\mathbf{I}$ and trained with Adam at a learning rate of $10^{-4}$, weight decay $10^{-4}$, a cosine schedule, mini-batches of $256$ for BAR dataset and $512$ for all other datasets, drawn from $\mathcal{P}\cup\mathcal{Q}$. The frozen encoder receives no gradient at any point. $\mathbf{W}$ is the only component the objective updates, at $d^2 \approx 0.59$M parameters against the backbone's ${\sim}300$M.

\paragraph{Classification.}
Under the cosine rule, we classify by similarity between the projected image embedding and the frozen class-wise prompt embeddings, exactly as CLIP-cosine does, and fit nothing. Under the linear head we fit a linear classifier (logistic regression) with balanced class weights on the projected training split, with $\mathbf{W}$ held fixed and no further tuning. The CLIP-linear reference uses an identical head on unrectified embeddings, so the two differ only in the presence of $\mathbf{W}$.

\paragraph{Hyperparameters.}
The two thresholds and margins are selected per dataset.
For the push set, $m_{\text{push}}$ is always chosen below $\tau_{\text{push}}$,
so the hinge pushes selected pairs past the threshold rather than merely toward
it.
For the pull set, $m_{\text{pull}}$ is always chosen above $\tau_{\text{pull}}$,
so the hinge pulls selected pairs beyond the threshold.
\Cref{subsec:supp_sensitivity} gives the swept grids and the configuration marked
on each grid is the one selected and used for the reported results.

\begin{algorithm}[t]
\caption{Proposed Approach}
\label{alg:method_supp}
\begin{algorithmic}[1]
\Require frozen embeddings $\{\mathbf{x}_i\}_{i=1}^{n}$, labels $\{y_i\}$,
         thresholds $\tau_{\text{push}}, \tau_{\text{pull}}$,
         margins $m_{\text{push}}, m_{\text{pull}}$, steps $T$
\Ensure  $\mathbf{W}$ and a linear head
\State $\mathcal{P} \gets \{(i,j) : i<j,\; y_i \neq y_j,\;
        \langle \mathbf{x}_i,\mathbf{x}_j\rangle > \tau_{\text{push}}\}$
\For{$i = 1,\dots,n$}
  \State $\mathcal{C}_i \gets \{j : y_j = y_i,\;
          \langle \mathbf{x}_i,\mathbf{x}_j\rangle < \tau_{\text{pull}}\}$
\EndFor
\State $\mathcal{Q} \gets \{(i,j) : \mathcal{C}_i \neq \varnothing,\;
        j \sim \mathrm{Unif}(\mathcal{C}_i)\}$,
        capped at $|\mathcal{P}|/N$ per class among the $N$ classes
\State $\mathbf{W} \gets \mathbf{I}_d$
\For{$t = 1,\dots,T$}
  \State draw a mini-batch $\mathcal{B} \subseteq \mathcal{P} \cup \mathcal{Q}$;\;
         $\mathbf{z}_i \gets \mathbf{W}\mathbf{x}_i /
         \lVert\mathbf{W}\mathbf{x}_i\rVert$
  \State $\mathbf{W} \gets \textsc{Adam}\big(\mathbf{W},
         \nabla_{\mathbf{W}}\mathcal{L}(\mathcal{B})\big)$
\EndFor
\State fit a linear head on $\{\mathbf{z}_i\}$ over the training split
\end{algorithmic}
\end{algorithm}



\section{Baselines}
\label{sec:supp_baselines}

Every method is run on the same frozen features and evaluated on the same split. For SEM, we use the bias-agnostic variant $\text{SEM}_i$, which names no spurious attribute, and its released implementation returns image embeddings unchanged. Therefore, for $\text{SEM}_i$, we cannot have any linear-head variant. For the methods that do transform image embeddings we obtain the head variant by fitting the logistic regression of \cref{sec:supp_impl} on their rectified features.

\subsection{Oracle specifications supplied to PRISM}
\label{subsec:supp_oracle}
 
The oracle experiment replaces the attributes that a language model proposes with an expert-written specification of the two shortcuts UrbanCars contains, leaving every other part of PRISM and PRISM-mini approaches unchanged. The specification names the background and the co-occurring object and enumerates the values each takes. The two methods consume it differently. PRISM generates scene descriptions from the specification and learns a projection under which descriptions differing only in an attribute become indistinguishable. Therefore, what it receives is a set of sentences. PRISM-mini trains nothing and directly builds its orthogonal projection from the attribute directions. Therefore, it receives only the values of the bias attributes, not descriptions.
 
The descriptions given to PRISM cover the eight cells of the class $\times$ background $\times$ co-occurring object \Cref{tab:supp_oracle_prompts}, so that the class and each attribute vary independently. Within a cell, incidental properties such as color, vehicle type, lighting, and road surface vary freely, and only the class and the two attributes are held to the values the cell specifies, which is what allows the objective to attribute the contrast to the attribute rather than to some correlated detail. One description per cell is shown below in \Cref{tab:supp_oracle_prompts} for demonstration purposes.  

\begin{table*}[t]
\centering
\small
\renewcommand{\arraystretch}{1.25}
\begin{tabular}{p{0.1\linewidth}p{0.1\linewidth}p{0.1\linewidth}p{0.60\linewidth}}
\toprule
\textbf{Class} & \textbf{Background} & \textbf{Co‑obj} & \textbf{Prompt} \\
\midrule
urban car & urban & urban &
\texttt{"A sleek black SUV cruises down a neon‑lit city avenue, passing a bright orange traffic cone near the curb."} \\
urban car & urban & country &
\texttt{"A sleek urban car is parked on a bustling downtown street while a curious cow grazes on a patch of grass beside the curb."} \\
urban car & country & urban &
\texttt{"A glossy urban car is parked on a dusty dirt road beside a bright red fire hydrant, with golden wheat fields stretching behind it."} \\
urban car & country & country &
\texttt{"A sleek urban car is parked on a grassy farm lane, while a curious cow grazes beside a weathered wooden barn in the background."} \\
country car & urban & urban &
\texttt{"A country car is parked on a bustling city street lined with brick buildings, next to a bright red fire hydrant."} \\
country car & urban & country &
\texttt{"A bright red country car is stopped at a busy city intersection, with a lone cow standing on the sidewalk beside it."} \\
country car & country & urban &
\texttt{"A dusty country car sits on a dirt road beside a bright red fire hydrant."} \\
country car & country & country &
\texttt{"A bright red country car rolls across a golden field as a curious cow watches from the edge of the grass."} \\
\bottomrule
\end{tabular}
\caption{\textbf{Oracle prompts used for UrbanCars.} Each row shows one prompt per combination of class, background, and co‑occurring object. These prompts replace the LLM‑generated descriptions in PRISM.}
\label{tab:supp_oracle_prompts}
\end{table*}

\section{Additional Results}
\label{sec:supp_results}

\subsection{Seed sweep}
\label{subsec:supp_seeds}

{\Cref{tab:seeds_rest} reports our results on BAR, BFFHQ, and Waterbirds datasets over three seeds under both classification rules. }


\begin{table}[t]
\centering
\small
\setlength{\tabcolsep}{2pt}
\renewcommand{\arraystretch}{1.12}
\resizebox{\columnwidth}{!}{
\begin{tabular}{l l cc cc}
\toprule
\textbf{Dataset} & \textbf{Ratio}
& \multicolumn{2}{c}{\textbf{Cosine}}
& \multicolumn{2}{c}{\textbf{Linear head}} \\
\cmidrule(lr){3-4}\cmidrule(lr){5-6}
& 
& \textbf{Ovr.} & \textbf{Worst}
& \textbf{Ovr.} & \textbf{Worst} \\
\midrule

BAR        & $1\%$    & $97.31_{\pm 0.30}$ & $87.79_{\pm 0.66}$ & $97.54_{\pm 0.18}$ & $83.10_{\pm 1.41}$ \\
\addlinespace[2pt]

BAR        & $5\%$    & $97.83_{\pm 0.22}$ & $87.79_{\pm 0.66}$ & $97.89_{\pm 0.00}$ & $84.51_{\pm 0.00}$ \\
\addlinespace[2pt]

BFFHQ      & $0.5\%$  & $95.03_{\pm 0.17}$ & $88.00_{\pm 0.57}$ & $92.67_{\pm 0.25}$ & $84.00_{\pm 0.40}$ \\
\addlinespace[2pt]

BFFHQ      & $1\%$    & $95.50_{\pm 0.51}$ & $90.80_{\pm 0.86}$ & $94.30_{\pm 0.17}$ & $86.67_{\pm 0.23}$ \\
\addlinespace[2pt]

BFFHQ      & $2\%$    & $96.60_{\pm 0.08}$ & $92.67_{\pm 0.19}$ & $95.60_{\pm 0.26}$ & $90.67_{\pm 0.46}$ \\
\addlinespace[2pt]

BFFHQ      & $5\%$    & $97.70_{\pm 0.08}$ & $95.47_{\pm 0.19}$ & $97.93_{\pm 0.12}$ & $95.87_{\pm 0.23}$ \\
\addlinespace[2pt]

Wbirds & 5\%       & $89.29_{\pm 0.02}$ & $81.95_{\pm 0.44}$ & $89.35_{\pm 0.07}$ & $80.01_{\pm 0.15}$ \\

\bottomrule

\end{tabular}
}
\caption{\textbf{Seed sweep on BAR, BFFHQ and Waterbirds datasets.}
Mean and standard deviation over three seeds.}
\label{tab:seeds_rest}
\end{table}

\subsection{Sensitivity to the selection thresholds}
\label{subsec:supp_sensitivity}

\begin{figure}[t]
\centering
\includegraphics[width=\linewidth]{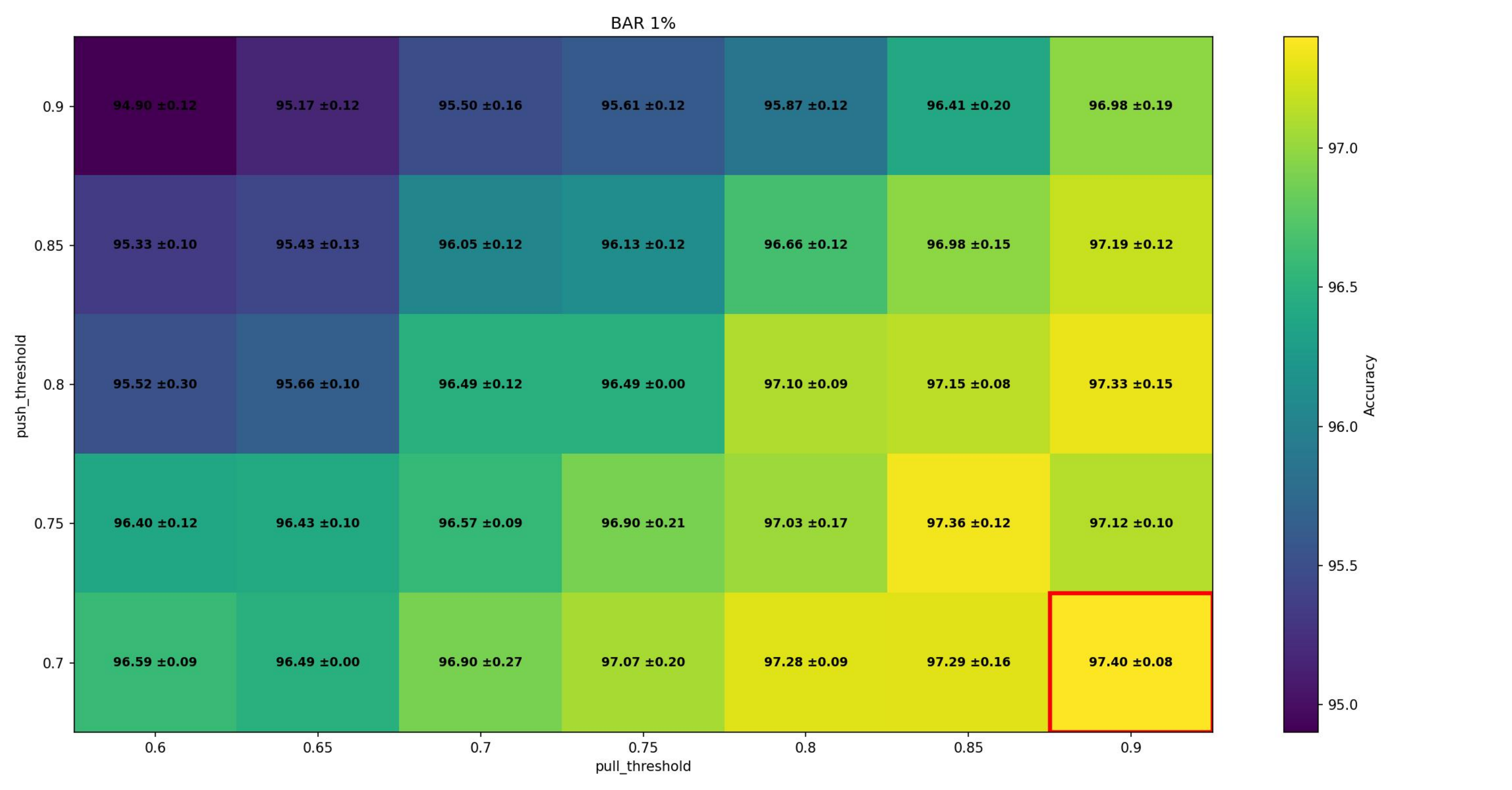}
\caption{\textbf{Threshold stability on BAR at $1\%$ conflicting ratio.}
Each cell reports the mean and spread across the four margin settings at that
threshold pair.
The red border box marks the selected configuration.}
\label{fig:supp_sensitivity_bar_1}
\end{figure}

\begin{figure}[t]
\centering
\includegraphics[width=\linewidth]{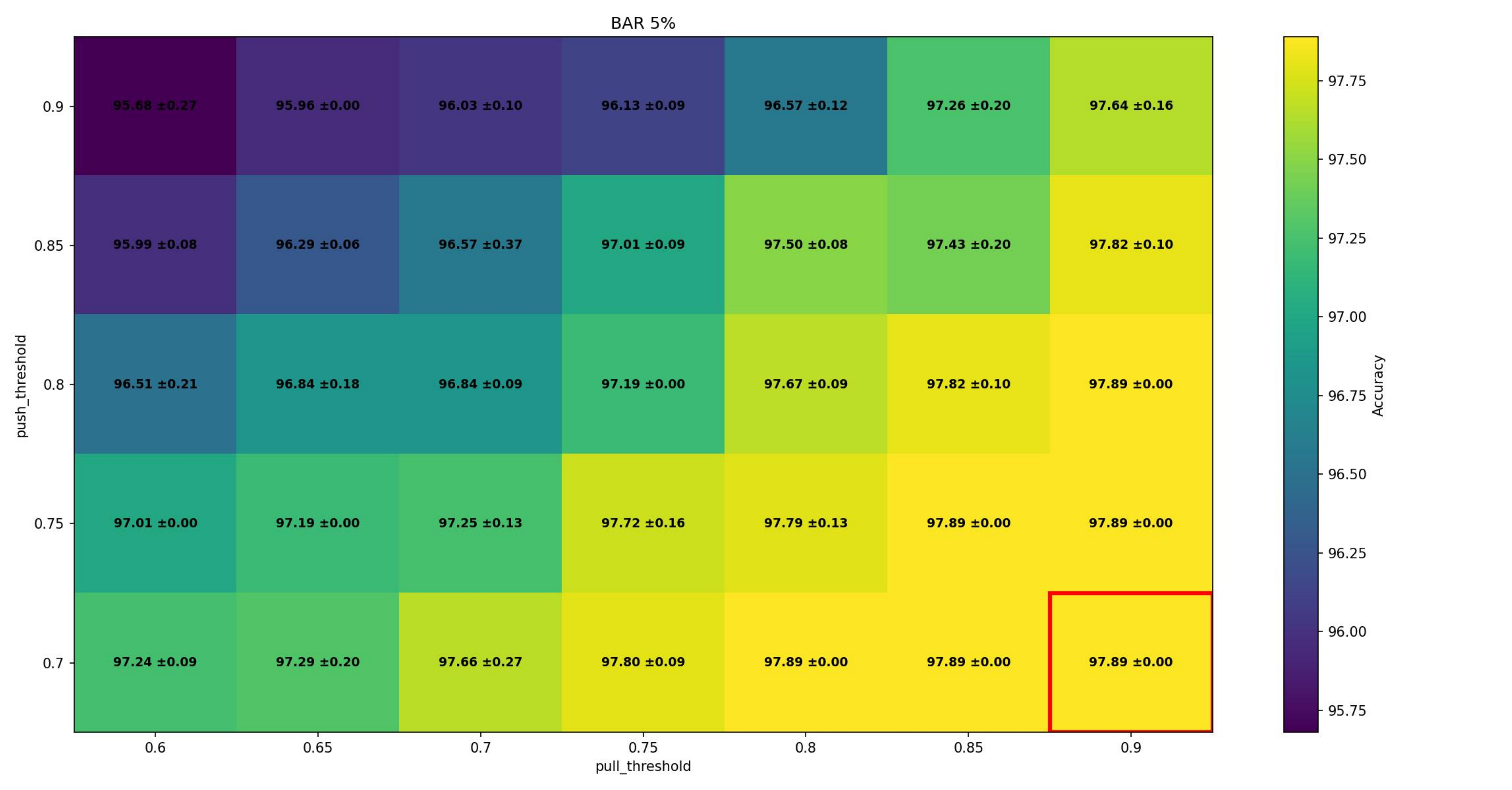}
\caption{\textbf{Threshold stability on BAR at $5\%$ conflicting ratio.}
Each cell reports the mean and spread across the four margin settings at that
threshold pair.
The red border box marks the selected configuration.}
\label{fig:supp_sensitivity_bar_5}
\end{figure}

\begin{figure}[t]
\centering
\includegraphics[width=\linewidth]{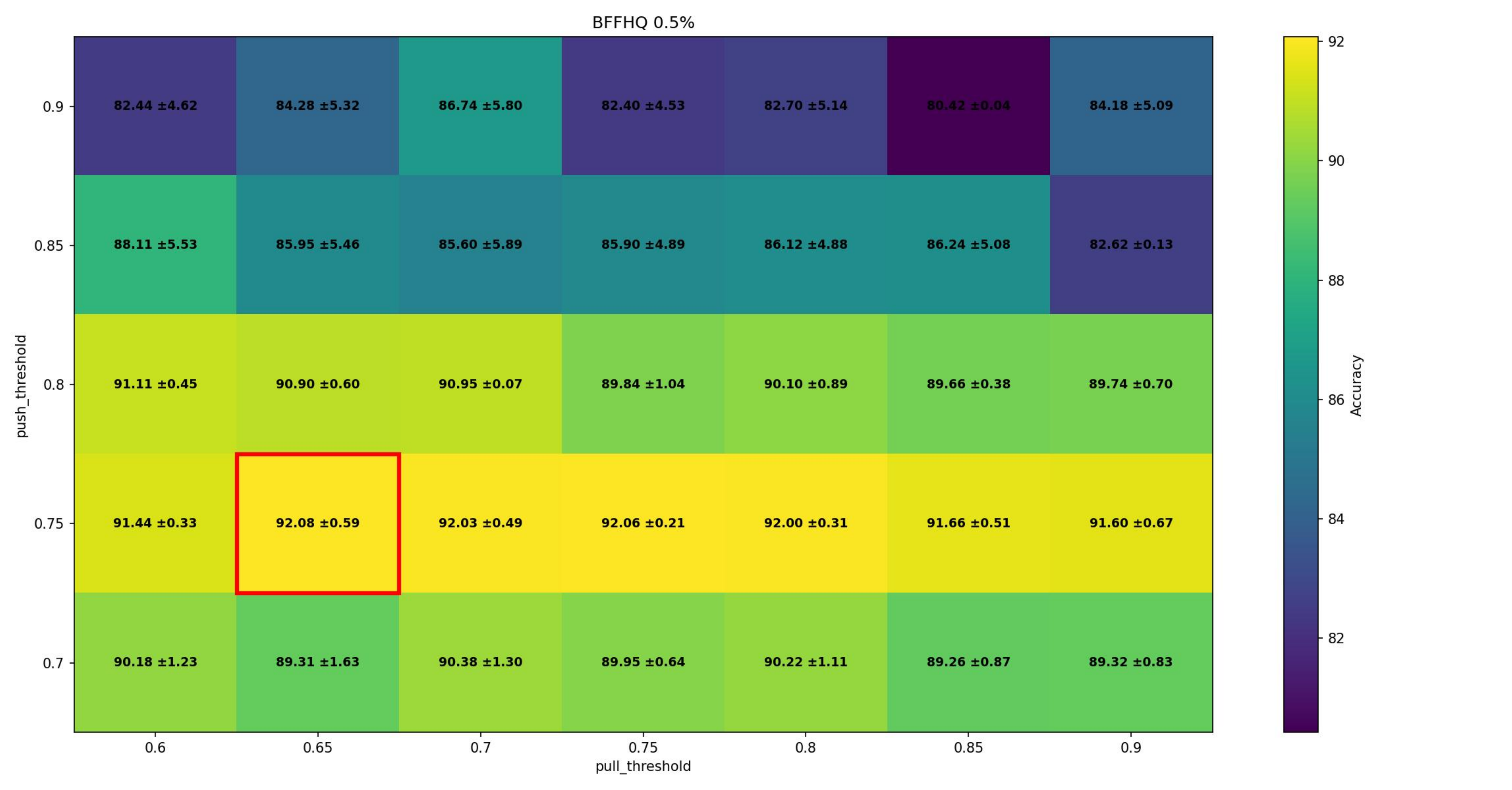}
\caption{\textbf{Threshold stability on BFFHQ at $0.5\%$ conflicting ratio.}
Each cell reports the mean and spread across the four margin settings at that
threshold pair.
The red border box marks the selected configuration.}
\label{fig:supp_sensitivity_bffhq_05}
\end{figure}

\begin{figure}[t]
\centering
\includegraphics[width=\linewidth]{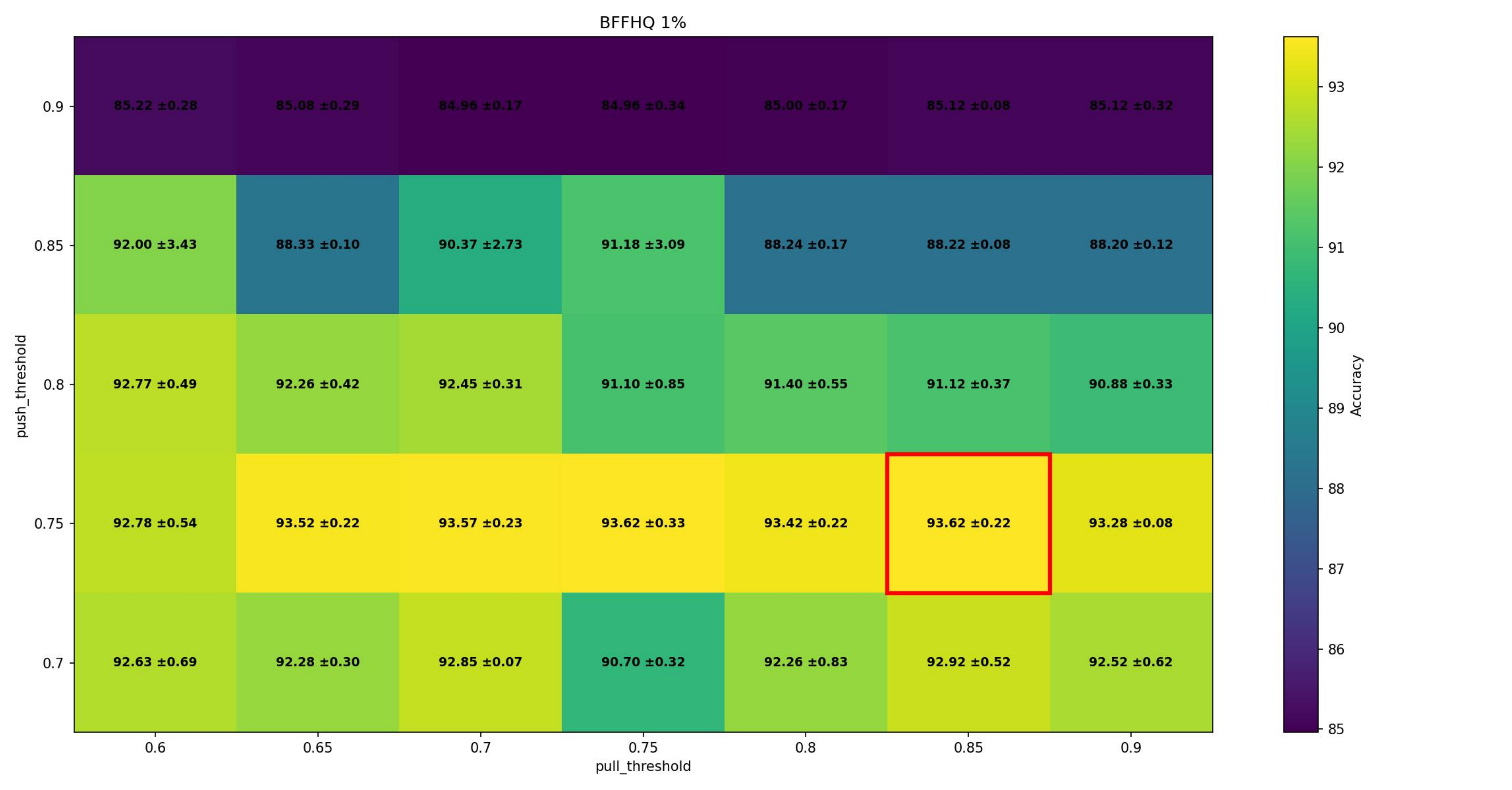}
\caption{\textbf{Threshold stability on BFFHQ at $1\%$ conflicting ratio.}
Each cell reports the mean and spread across the four margin settings at that
threshold pair.
The red border box marks the selected configuration.}
\label{fig:supp_sensitivity_bffhq_1}
\end{figure}

\begin{figure}[t]
\centering
\includegraphics[width=\linewidth]{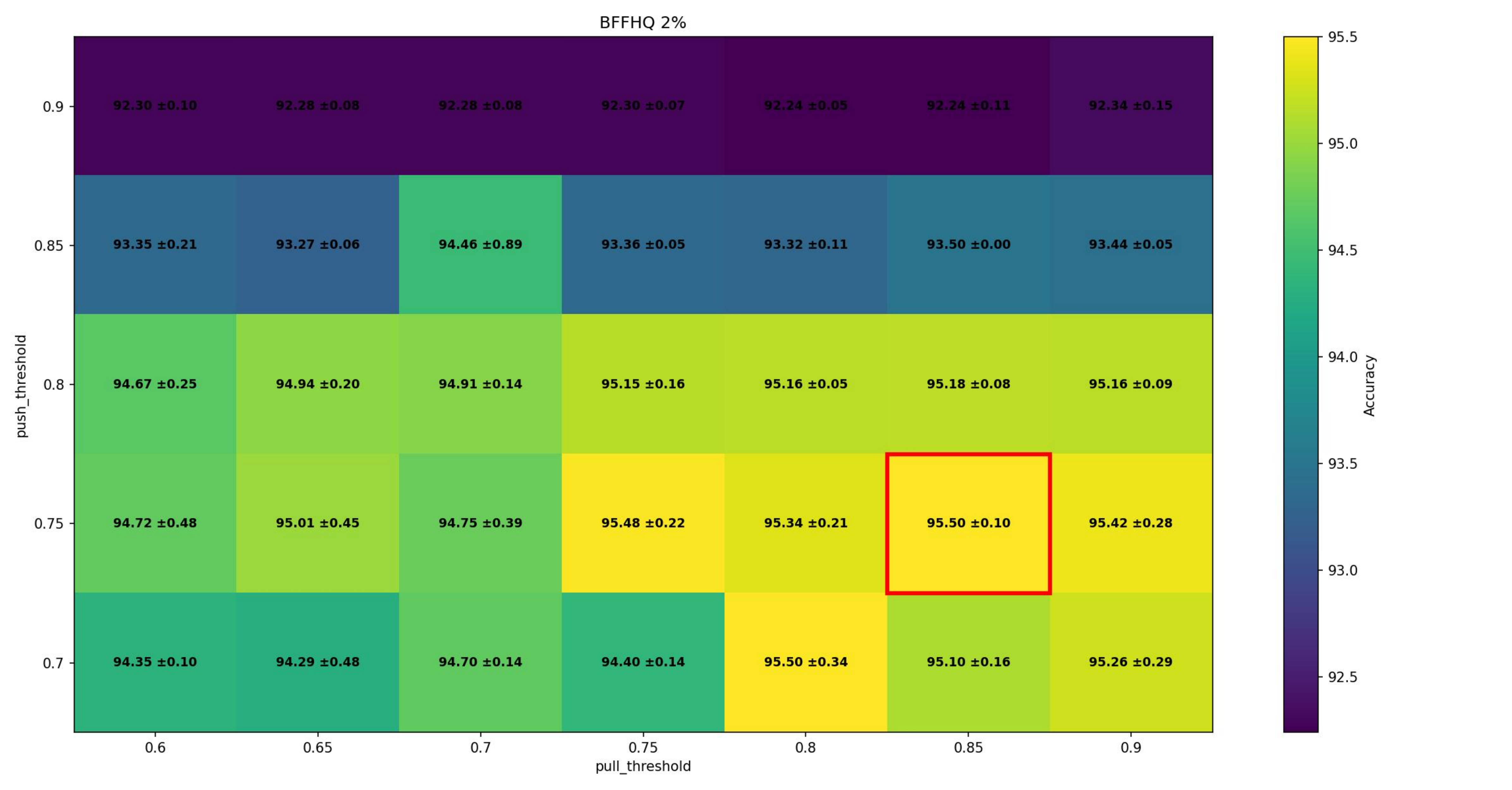}
\caption{\textbf{Threshold stability on BFFHQ at $2\%$ conflicting ratio.}
Each cell reports the mean and spread across the four margin settings at that
threshold pair.
The red border box marks the selected configuration.}
\label{fig:supp_sensitivity_bffhq_2}
\end{figure}

\begin{figure}[t]
\centering
\includegraphics[width=\linewidth]{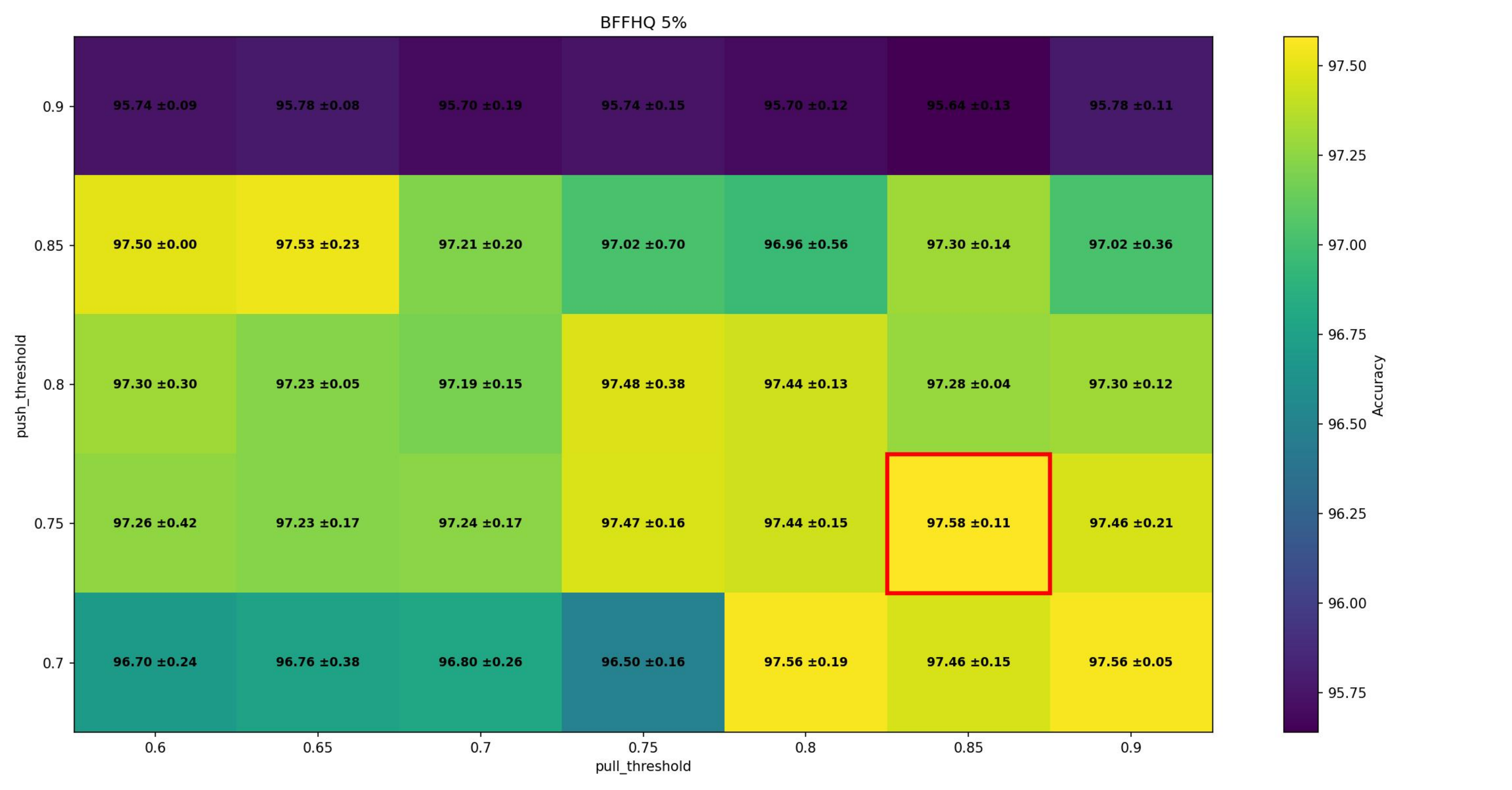}
\caption{\textbf{Threshold stability on BFFHQ at $5\%$ conflicting ratio.}
Each cell reports the mean and spread across the four margin settings at that
threshold pair.
The red border box marks the selected configuration.}
\label{fig:supp_sensitivity_bffhq_5}
\end{figure}

\begin{figure}[t]
\centering
\includegraphics[width=\linewidth]{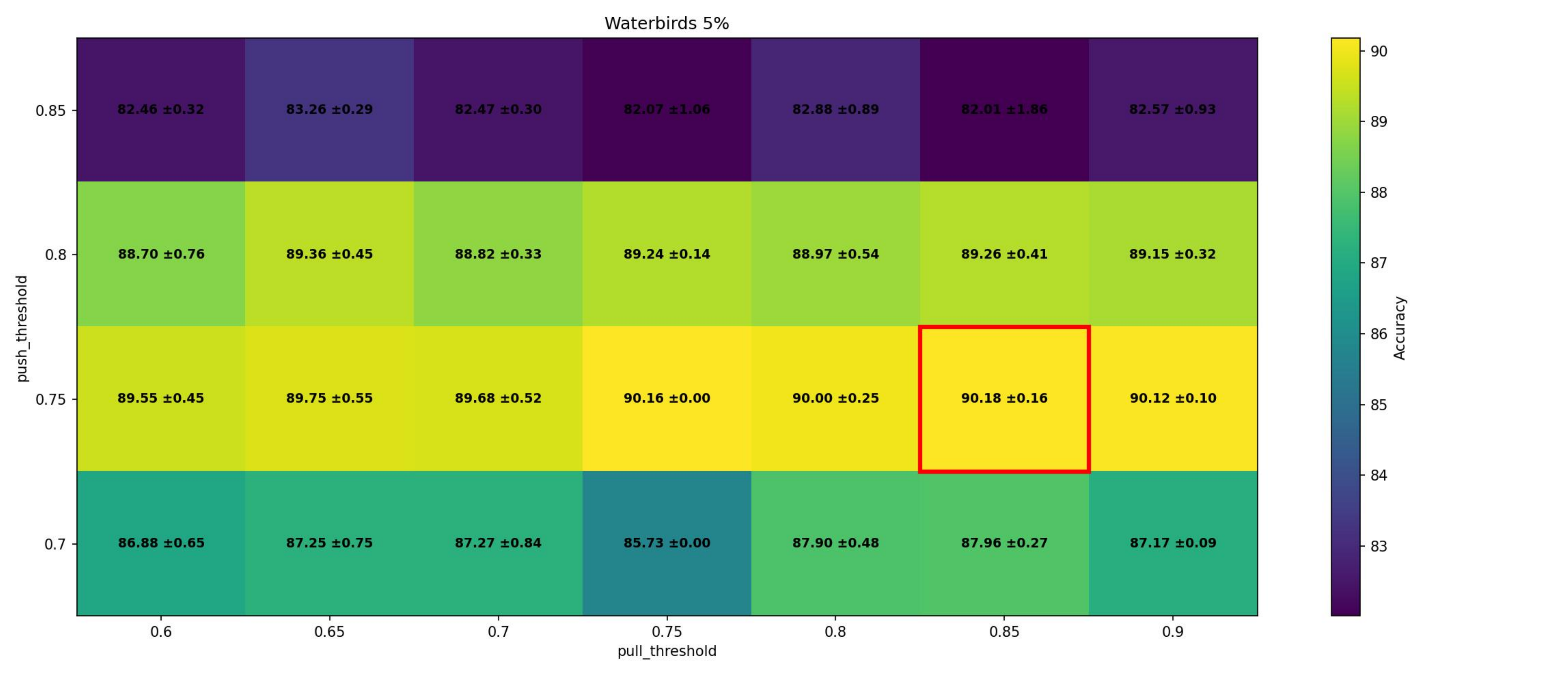}
\caption{\textbf{Threshold stability on Waterbirds.}
Each cell reports the mean and spread across the four margin settings at that
threshold pair.
The red border box marks the selected configuration.}
\label{fig:supp_sensitivity_waterbirds}
\end{figure}

\Cref{fig:supp_sensitivity_bar_1,fig:supp_sensitivity_bar_5,fig:supp_sensitivity_bffhq_05,fig:supp_sensitivity_bffhq_1,fig:supp_sensitivity_bffhq_2,fig:supp_sensitivity_bffhq_5,fig:supp_sensitivity_waterbirds}
show how accuracy varies with the two selection thresholds.
Each cell reports the mean and spread across the four margin settings at that threshold pair. The red border box marks the configuration selected and used for all reported results. For every dataset and ratio, a broad region around the selected cell performs within a few points of the best configuration, and the selected cell lies
within that region. Variance concentrates where the threshold has very few pairs to work with.


\subsection{Other backbones}
\label{subsec:supp_backbones}

Our method is not tied to a particular backbone. The objective computes cosine similarities between image embeddings. Therefore, it applies to any frozen encoder of a large pre-trained model.
\Cref{tab:backbones} repeats the UrbanCars experiment on SigLIP and DINOv3.
Accuracy improves on all four metrics for every backbone.
The improvement narrows as the frozen representation becomes less entangled to begin with.
On CLIP, Worst Joint accuracy rises by $45.16$ points, on SigLIP by $24.20$ points, and on DINOv3 by $3.22$ points.
These rows are single runs at a fixed seed, not the three means reported in the main paper.

The DINOv3 rows are possible only for our method.
Every frozen-backbone method we compare against derives its correction from
prompts, whether written by a practitioner, proposed by a language model, and DINO being a vision-only encoder supplies none. Our objective reads image-image similarity, so the text side is never consulted.

\begin{table}[t]
\centering
\small
\setlength{\tabcolsep}{1pt}
\renewcommand{\arraystretch}{1.15}
\begin{tabular}{l l cccc}
\toprule
\textbf{Backbone} & &
\textbf{Ovr.} & \textbf{W-BG} & \textbf{W-Co} & \textbf{W-Joint} \\
\midrule
CLIP ViT-L/14 & no debiasing & 83.87 & 68.55 & 65.32 & 38.71 \\
\rowcolor{oursrow}
              & $+$ ours     & 90.93 & 86.29 & 85.48 & 83.87 \\
\midrule
SigLIP ViT-L/16 & no debiasing & 87.70 & 79.03 & 78.23 & 62.90 \\
\rowcolor{oursrow}
                & $+$ ours     & 92.54 & 87.90 & 88.71 & 87.10 \\
\midrule
DINOv3 ViT-L/16$^{\dagger}$ & no debiasing & 92.34 & 86.29 & 87.90 & 80.65 \\
\rowcolor{oursrow}
                & $+$ ours     & 93.15 & 89.52 & 89.52 & 83.87 \\
\bottomrule
\end{tabular}
\caption{\textbf{The method transfers across frozen encoders.}
UrbanCars, with and without our projection, evaluated with the linear head.
Accuracy improves on all four metrics for every backbone, by a margin that
narrows as the undebiased representation improves.
Rows are single runs at a fixed seed.
$^{\dagger}$DINOv3 is a vision-only encoder and produces no text embeddings, so
none of the prompt-based methods we compare against can be applied to it and our method
does not use the text side.}
\label{tab:backbones}
\end{table}

\subsection{Class imbalance}
\label{subsec:supp_imbalance}

The benchmarks of the main paper are class-balanced, which is not typical of
data collected in the wild.
\Cref{tab:imbalance} resamples the UrbanCars training split so that the two
classes appear in a $70{:}30$ and a $90{:}10$ ratio, leaving the shortcut
correlations unchanged. Under both classification rules, our method performs significantly better than the undebiased reference of $38.71$. Therefore, the proposed method does not depend on the balanced construction of these benchmarks. 

\begin{table}[t]
\centering
\small
\setlength{\tabcolsep}{3pt}
\renewcommand{\arraystretch}{1.15}
\begin{tabular}{l l cccc}
\toprule
\textbf{Class ratio} & &
\textbf{Ovr.} & \textbf{W-BG} & \textbf{W-Co} & \textbf{W-Joint} \\
\midrule
\rowcolor{refrow}
\multicolumn{2}{l}{CLIP-cosine} & 50.00 &  4.03 & 13.71 &  0.00 \\
\rowcolor{refrow}
\multicolumn{2}{l}{CLIP-linear} & 83.87 & 68.55 & 65.32 & 38.71 \\
\midrule
\rowcolor{oursrow}
balanced & cosine & 89.11 & 76.61 & 76.61 & 66.13 \\
\rowcolor{oursrow}
         & head   & 90.93 & 86.29 & 85.48 & \best{83.87} \\
\midrule
\rowcolor{oursrow}
$70{:}30$ & cosine & \best{92.90} & \best{88.70} & \best{88.70} & \best{87.10} \\
\rowcolor{oursrow}
          & head   & 92.54 & 86.29 & 83.87 & 80.65 \\
\midrule
\rowcolor{oursrow}
$90{:}10$ & cosine & 88.30 & 80.60 & 75.80 & 66.10 \\
\rowcolor{oursrow}
          & head   & 90.50 & 87.90 & 80.60 & 80.60 \\
\bottomrule
\end{tabular}
\caption{\textbf{Robustness to class imbalance on UrbanCars.}
The training split is resampled so that the two classes appear in the stated
ratio, leaving the shortcut correlations unchanged.
Worst Joint accuracy never falls below twice the undebiased reference of $38.71$. Both classification rules are reported, and rows are single runs at a fixed
seed.}
\label{tab:imbalance}
\end{table}
\section{Waterbirds Evaluation Protocol}
\label{sec:supp_waterbirds}

For Waterbirds, we follow the evaluation protocol of \cite{sem}, including the
ViT-L/14@336px backbone. Baselines other than CLIP-linear and our own are reproduced using the \cite{sem} implementation. We train our method using the hyperparameter search procedure described in \cref{sec:supp_impl}.
The frozen encoder is CLIP ViT-L/14@336px. We apply $\mathbf{W}$ to the resulting embeddings and train the linear head on the rectified features.

\section{Additional Analysis}
\label{sec:supp_analysis}

\subsection{Class structure in the projected space}
\label{subsec:supp_separation}

\begin{figure}[t]
\centering
\includegraphics[width=\linewidth]{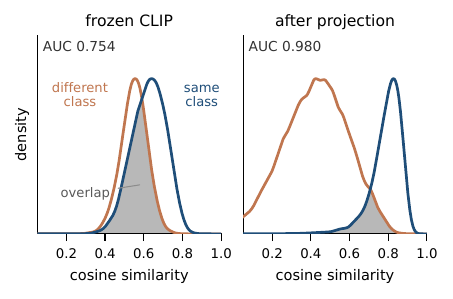}
\caption{\textbf{Class separation before and after the projection.}
Pairwise cosine similarity over UrbanCars training pairs, split by whether the
pair shares a class, with the overlap between the two shaded.
AUC is the probability that a randomly drawn same-class pair is more similar
than a randomly drawn cross-class pair and at $0.5$ similarity would carry no
information about class membership.
Curves are normalized to equal height, as same-class pairs are the more
numerous.}
\label{fig:separation}
\end{figure}

The main paper finds that in the frozen encoder a change of background costs
almost as much similarity as a change of class.
\Cref{fig:separation} repeats that measurement after projection.
Mean similarity rises from $0.626$ to $0.792$ for same-class pairs and falls
from $0.551$ to $0.421$ for cross-class pairs, widening the gap between them
from $0.075$ to $0.371$.
The fraction of cross-class pairs more similar than the average same-class pair
falls from $12.7\%$ to $0.45\%$, and the probability that a randomly drawn
same-class pair is more similar than a randomly drawn cross-class pair rises
from $0.754$ to $0.980$.
These are measurements on the training split and describe how the projection
reshapes the representation rather than adding evidence of generalization.

\subsection{Where conflicting samples move}
\label{subsec:supp_drift}

\begin{figure}[t]
\centering
\includegraphics[width=\linewidth]{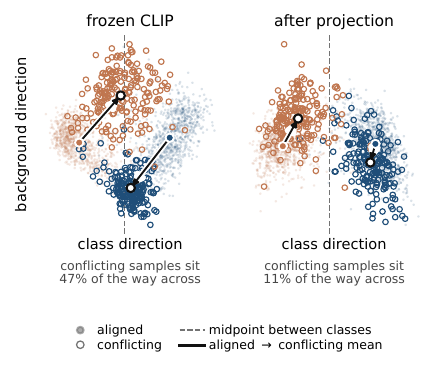}
\caption{\textbf{Where conflicting samples sit.}
UrbanCars training embeddings projected onto the plane spanned by the class
direction and the background direction, before and after our projection.
Filled points are bias-aligned samples and open points are bias-conflicting. Each
arrow runs from the aligned mean of a class to its conflicting mean, and the
dashed line marks the midpoint between the two class means.
Attribute labels are used for this measurement only and never during training.}
\label{fig:embedding}
\end{figure}

The main paper showed that baseline finds it difficult to identify bias-conflicting samples relative to the class boundary. Projecting the embeddings onto the plane spanned by the class direction and the background direction makes that position visible (\cref{fig:embedding}). Both directions are differences of group means computed after training, and the background direction is orthogonalized against the class direction so that the two axes are not redundant, and neither enters the objective. In the frozen space, a conflicting sample sits on average $47\%$ of the way toward the opposing class along the class direction. Therefore, many of these samples shift to the wrong side of a boundary fitted to the majority. After projection, the same measurement gives $11\%$. The plane retains two of $768$ dimensions and is shown for interpretation, and the similarity measurements above are computed in the full space.

{
    \small
    \bibliographystyle{ieeenat_fullname}
    \bibliography{main}
}